\documentclass[10pt,final,journal,letterpaper,twoside,twocolumn]{IEEEtran}

\usepackage{pgfgantt}
\usepackage{cite}
\usepackage{bm}
\usepackage{mathtools}
\DeclarePairedDelimiter{\abs}{\lvert}{\rvert}
\DeclarePairedDelimiter\norm{\lVert}{\rVert}%
\usepackage{amsmath} 
\usepackage{amssymb}  

\usepackage{cleveref}
\usepackage{breqn}
\usepackage{epstopdf}
\usepackage{algorithm}
\usepackage{algorithmicx}
\usepackage[cal=cm]{mathalfa}

\usepackage{algorithm}
\usepackage[noend]{algpseudocode}
\usepackage{varwidth}
\usepackage{array}
\usepackage{url}
\usepackage{float}
\usepackage{booktabs}
\usepackage{cite}
\usepackage{graphicx}
\usepackage{subfig}
\usepackage{amssymb}
\usepackage[english]{babel}
\usepackage{stmaryrd}

\newcolumntype{L}[1]{>{\raggedright\let\newline\\\arraybackslash\hspace{0pt}}m{#1}}
\newcolumntype{C}[1]{>{\centering\let\newline\\\arraybackslash\hspace{0pt}}m{#1}}
\newcolumntype{R}[1]{>{\raggedleft\let\newline\\\arraybackslash\hspace{0pt}}m{#1}}

\makeatletter
\let\oldabs\abs
\def\abs{\@ifstar{\oldabs}{\oldabs*}}
\let\oldnorm\norm
\def\norm{\@ifstar{\oldnorm}{\oldnorm*}}
\makeatother

\begin{document}
\title{Time-to-Green predictions for fully-actuated signal control systems with supervised learning}

\author{Alexander Genser,  Michail A. Makridis, Kaidi Yang, Lukas Ambühl, Monica Menendez and Anastasios Kouvelas
}
\maketitle
\begin{abstract}
Recently, efforts have been made to standardize signal phase and timing (SPaT) messages. These messages contain signal phase timings of all signalized intersection approaches. This information can thus be used for efficient motion planning, resulting in more homogeneous traffic flows and uniform speed profiles. Despite efforts to provide robust predictions for semi-actuated signal control systems, predicting signal phase timings for fully-actuated controls remains challenging. 
This paper proposes a time series prediction framework using aggregated traffic signal and loop detector data. We utilize state-of-the-art machine learning models to predict future signal phases' duration. The performance of a Linear Regression (LR), a Random Forest (RF), and a Long-Short-Term-Memory (LSTM) neural network are assessed against a naive baseline model. Results based on an empirical data set from a fully-actuated signal control system in Zurich, Switzerland, show that machine learning models outperform conventional prediction methods. Furthermore, tree-based decision models such as the RF perform best with an accuracy that meets requirements for practical applications.  

\end{abstract}
\section{INTRODUCTION}
\label{sec:introduction}
Digitization has substantially transformed the transportation sector over the past decade. The availability of several new data sources (e.g., sensor and in-vehicle technologies) enables data-driven methods to be integrated into established traffic management systems. In addition, new developments such as vehicle-to-infrastructure (V2I) communications open the possibility for new methodologies that utilize infrastructure data for motion planning, speed advisory systems, or route choice~\cite{ref:claussmann_motion_planning}. Recent developments in traffic signal control at urban intersections (e.g., fully-actuated signal control~\cite{ref:ZHENG2013_actuated_SC}, self-steering algorithms~\cite{ref:laemmer_2008}) affect signal phasing and result in different green, red, and cycle times. Therefore, it would benefit speed advisory systems if the duration of a future signal phase is known.
Ideally, fewer vehicles have to stop when crossing an intersection and uncertainty for other transportation modes is reduced. Signal phasing and timing (SPaT) messages provide the necessary information. Unfortunately, determining the future phase duration of fully-actuated signal control systems is not trivial reverse engineering as predictions depend on Loop Detector (LD) detections that occur after the forecast is applied. Also, such systems typically involve complex optimization, which constitutes a barrier to applying SPaT messages in practice. Therefore, a sophisticated modeling approach using traffic signals and LD data for accurate predictions is still a subject of research.

In this paper, we propose a methodology to forecast the duration of the following red phase (i.e., when a traffic stream is  not allowed to cross the intersection and a stop is required). By providing an accurate prediction for the next red-phase (in this work denoted as the Time-to-Green (T2G)) with Machine Learning (ML) models, we can enhance SPaT messages. We utilize empirical traffic signal and LD data based on previous work~\cite{ref:genser_t2g_itsc}, and compute domain-specific features for time series forecasting. We first introduce a simple reference model: ``no-change'', a dummy model that utilizes the duration of the last occurring red phase to justify the use of ML. Then, a selection of complex models found to be strong candidates for various ML problems are compared against the dummy case. Finally, we implement a Linear Regression (LR), a Random Forest (RF) regressor, and a Long-Short-Term-Memory (LSTM) neural network.  The proposed framework allows for (a) the processing of empirical traffic signal and LD data, (b) extensive feature engineering, and (c) the assessment of supervised machine learning models' phase predictions. A numerical experiment in Zurich, Switzerland, is conducted to prove the concept. A data set from a fully-actuated signal control system, consisting of historical LD and traffic signal data, is utilized. The area under investigation also includes a priority for public transportation (i.e., signal priorities frequently change the control behavior of the intersection), which further demonstrates the complexity of the problem.

An accurate prediction of the T2G can not only help the improvement of speed-advisory systems but consequently also have an impact on the homogeneity of traffic flow in multi-modal urban transportation networks. We address the opening challenges for T2G predictions by providing the following contributions:  

\begin{enumerate}
    \item The framework design allows for predicting the next signal phase of multi-modal fully-actuated signal control systems. The work is based on an empirical data set allowing for real-time applications. 

    \item The prediction of the next red phase, modeled as a supervised learning problem, captures the complex and non-linear relation between a traffic signal and LD detection data. 
   
    \item The framework requires no prior knowledge about the implemented traffic signal control system. Hence, the method also provides accurate predictions where the signal control algorithm is proprietary.
   
    \item A physics-informed feature engineering incorporates the concepts of traffic flow theory. The approach enhances the quality of a given prediction model and can be used for multi-model systems with transit priority.
   
\end{enumerate}

The remainder of this paper is organized as follows: Section~\ref{sec:back_motivation} provides an overview of recent research on the prediction of signal phasing and timing. Besides, due to the limited contributions in this area utilizing ML techniques, we provide an overview of publications applying such techniques to similar transportation problems. In Section~\ref{sec:defintion} the time series problem is defined. Section~\ref{sec:methodology} describes the utilized data sources and the feature engineering based on LD and traffic signal data. Furthermore, the framework definition and the theory of the selected models is introduced. Finally, we introduce the performance metrics used to evaluate the T2G predictions. Section~\ref{sec:num_exp} shows the applicability of all models based on a case study with a detailed presentation of prediction results and a final performance evaluation. Finally, a discussion, conclusion, and proposal of future work are given in Section~\ref{sec:diss} and Section~\ref{sec:diss_concl}. 

\section{RELATED WORKS}
\label{sec:back_motivation}
Recently, efforts were made to standardize SPaT messages~\cite{ref:c2c}. Such messages contain the current phase with a prediction for the corresponding phase duration for all signalized intersection approaches. Hence, SPaT information allows a more efficient and environmentally friendly motion planning of human-driven and/or autonomously operated individual or public transport vehicles. Especially in urban areas, this would lead to more homogeneous traffic flow, a smoother speed profile (i.e., the absence of speeding and heavy breaking between traffic lights) or an improvement in ride comfort~\cite{ref:genser_comfort_cacaie}. In this section, we first present methods for obtaining SPAT estimations/predictions (Section~\ref{sec:RL_SPAT}) and continue with related works that specifically focus on transportation problems with ML applications (Section~\ref{sec:RL_ML}).

\subsection{Prediction methods for SPaT information}
\label{sec:RL_SPAT}
Most of the existing methods to obtain SPaT information for semi-actuated traffic signal control systems are based on aggregated trajectory data. In such methods, signal timings are unknown and can either be fixed or change slowly in time. They use estimation approaches based on floating car data~\cite{ref:fayazi1, ref:fayazi2, ref:wang, ref:Yu} or travel time measurements collected with wireless traffic sensors~\cite{ref:ban}.
For example,~\cite{ref:fayazi1} and~\cite{ref:fayazi2} employed a queue discharging model to estimate the start of green signals based on aggregated low-frequency bus data and probe data. Yu et al.~\cite{ref:Yu} formulated the SPaT estimation problem into a general approximate greatest common divisor problem, aiming to obtain the cycle lengths, green times, and the phase schemes based on historical sparse taxi trajectories. Protschky et al.~\cite{ref:protschky} used a Bayesian learning approach to reconstruct the cycle length from historical trajectory data for traffic signals where the cycle length is fixed within a certain period. These methods typically rely on the underlying assumption that the cycle length is fixed, although some of them (e.g.,~\cite{ref:fayazi2,ref:protschky}) are able to identify the occasional changes in the traffic signal timing plan. Moreover, these works are based on aggregating historical vehicle trajectories, assuming that the historical signal timings are unknown. They do not provide insights on how the spatially-sparse LD data and the temporally-sparse public transport data can be utilized to perform real-time SPaT prediction within each signal cycle. Therefore, their applicability to real-life problems is limited.

Some other studies propose probabilistic methods~\cite{ref:protschky3, ref:Ibrahim, ref:zhu} and ML techniques~\cite{ref:protschky2, ref:islam_cnn_lstm_spat, ref:genser_t2g_itsc} that can be used to predict the SPaT information for actuated and adaptive traffic signals. Compared to the estimation problems above, here traffic signal data (i.e., signal timings) are available based on a historical data set. Protschky et al.~\cite{ref:protschky2} employed a Kalman Filter (KF) to estimate the probability of phase switches at each time step using historical traffic signal data. This work was further enhanced to consider implementation factors such as latency and data losses~\cite{ref:protschky3}. Based on historical signal data,~\cite{ref:Ibrahim} estimated the conditional distribution of each signal phase given the real-time signal phase measurements and predicted the phase duration as the conditional expectation and the confidence interval. These methods treat the SPaT information as a time series and are expected to yield satisfactory prediction accuracy if the variance of the signal phase duration is small. However, in cases where the signal phase duration changes drastically (i.e., with high variance), these works may not yield the best results, as they cannot incorporate relevant vehicle detection information. ~\cite{ref:zhu} took an initial step to establish the relationship between real-time vehicle information and traffic signal timings. Based on historical floating car and bus trajectory data, it first calculated the joint distribution of the driving speed and the distance to the stop line, given a particular signal state (green or red). Afterward, it predicted the phase duration using a maximum a posteriori (MAP) estimation. This work only linked the signal state to the information of an individual vehicle at a single time step. However, in reality, many detectors can contribute to the signal timings in a complex signalized intersection with multiple approaches and movements. Hence, learning the relationship between the signal states and the information sent by multiple detectors is crucial as shown in~\cite{ref:islam_cnn_lstm_spat}. Finally, previous work from the authors in~\cite{ref:genser_t2g_itsc} shows a preliminary ML approach with traffic signal and LD data. Nevertheless, the non-aggregated raw data is utilized for predicting the T2G without extensive feature engineering. Also, the set of ML models is tested on a small data set, leading to an overoptimistic result for the LSTM neural networks. 

\subsection{Machine learning based methods for similar problems in transportation}
\label{sec:RL_ML}
Despite the lack of literature on ML-based SPaT prediction for actuated and
adaptive signals, ML-based methods have been widely applied to many transportation research topics. Here, we present a short literature review on ML applications on similar prediction problems. Interested readers can refer to~\cite{ref:Wang_LSTM, ref:zhang} for comprehensive surveys. Two important attributes characterize the SPaT prediction problem: First, it is a prediction problem aiming to obtain a future traffic signal state using historical data. Due to the uncertainty of the future arrivals (thus the actuation of the detectors), there is uncertainty in the future signal state. Second, it aims to establish the relationship between the traffic signal state and detector information. Therefore, in this subsection, we mainly look at two research problems, the short-term traffic prediction that shares the first attribute and the prediction of driver behaviors (e.g., acceleration
rates, lane change decisions) that shares the second.

Within the same family of problems, we can find the short-term prediction of traffic variables, including traffic states (e.g., flow, speed, occupancy), demand (e.g., origin-destination matrix), and accident rates. Such problems are typically formulated as a time-series prediction problem where future variables are predicted from historical ones. Conventional parametric methods, such as Auto-Regressive Integrated Moving Average (ARIMA) and KF, can achieve good performance when the traffic variations are regular. To handle more general traffic scenarios, many ML models have been adopted, such as k-Nearest Neighbor (k-NN)~\cite{ref:chang}, multivariate regression~\cite{ref:sun, ref:clark, ref:genser}, Support Vector Regression (SVR)~\cite{ref:hong, ref:castro_neto, ref:jeong}, RF~\cite{ref:zhang_RSF}, artificial neural networks (ANN)~\cite{ref:vlahogianni, ref:chan, ref:kumar, ref:tang}, and deep learning methods~\cite{ref:lv, ref:huang, ref:yu_rose, ref:qian}. It is non-trivial to compare the performance of the proposed methods as these methods are developed and evaluated based on different data sets with specific features. Nevertheless, results indicate that the deep neural networks can outperform other ML methods with sufficient training data~\cite{ref:yu_rose, ref:koesdwiady}. 

Another traffic problem that shares some similarities with the SPaT prediction problem is the prediction of driver behaviors. This problem links the behaviors of the drivers (e.g., acceleration
rates, lane change decisions) with the current traffic scenarios (e.g., position and speed of
the considered vehicle and the vehicles around it). In addition to the traditional analytical car
following and lane changing models~\cite{ref:panwai, ref:ciuffo, ref:kesting}, many works attempt to employ data driven models to capture driver behaviors based on methods such as Hidden Markov Models~\cite{ref:tran}, support vector machines (SVM)~\cite{ref:kim, ref:kumar_2}, Bayesian Filter~\cite{ref:kumar_2}, etc. Deep learning methods have also attracted much attention within the context of this research problem. For example,~\cite{ref:gurghian} employed a deep
Convolutional Neural Network (CNN) to perform lane change prediction based on camera data;~\cite{ref:oluwatobi} predicted the actions of drivers using deep Recurrent Neural Network (RNN) based on in-vehicle sensors;~\cite{ref:wang_car_following} modeled car-following behaviors by deep RNN with the Gated Recurrent Unit (GRU) using the position and speed information over multiple time steps. However, these methods typically rely on high-resolution and demanding data sets (e.g., GPS data or in-vehicle sensors).

This paper focuses on machine learning methods that use traffic signals and data from loop detectors, the most common traffic data source in cities worldwide. Our work proposes the first framework to provide a robust prediction of the next signal phase in a multi-modal and fully-actuated control system with public transportation priority. Furthermore, our work captures complex non-linear relationships between a traffic signal and detector data by applying physics-informed feature engineering. 

\section{PROBLEM DEFINITION}
\label{sec:defintion} 
Assume an intersection controlled by fully-actuated signal control with installed traffic lights and LDs. There are $A$ traffic lights and $B$ LDs, and historical states are available from all devices (i.e, a traffic light operated in a red or green phase; an LD occupied/not occupied). We denote every traffic light with the index $i\in \mathcal{S}$, where $\mathcal{S} = \{ 1, 2, 3, \dots, A \}$ and every LD with the index $j\in \mathcal{D}$, where $\mathcal{D} = \{ 1, 2, 3, \dots, B \}$. As fully-actuated signal controls allow for non-constant red and green times, it is essential to distinguish traffic lights by the index $i$. Hereafter, we define a signal cycle as $c_{i,n}$. $n$ denotes the index of a specific cycle $c$ of a traffic light $i$. Every $c_{i,n}$ starts with a red phase for $i$ and ends when the following red phase starts. Because we consider a fully-actuated control, our definition implies that different signals $i$ from the same intersection might have different cycles at the same time.  

The signal states of all traffic lights and LDs during the corresponding cycle $c_{i,n}$ are then utilized to compute a feature set \textbf{X}$_n$.
Aiming for the prediction of the T2G, denoted as $\hat{Y}_{i}(c_{i,n})$, the problem is formulated as $\hat{Y}_{i}(c_{i,n}) = f(\textbf{X}_n)$. The function $f(\textbf{X}_n)$ denotes the relationship between the set of input features $\textbf{X}_n$ and the target $\hat{Y}_{i}(c_{i,n})$. Note that the T2G prediction corresponds to the red time of the next cycle $c_{i,n+1}$. 

\section{METHODOLOGY}
\label{sec:methodology}
\subsection{T2G framework}
\label{sec:framework}
\begin{figure*}[!t]
    \centering
    \includegraphics[width=1\textwidth]{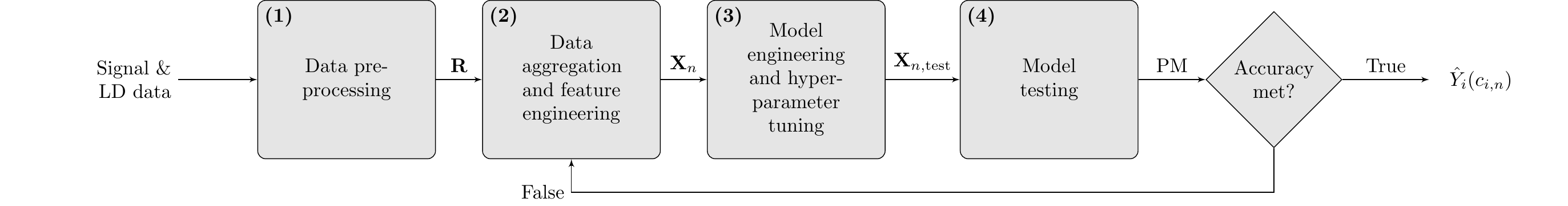}
    \caption{T2G framework. The input data includes by LD and signal data from the traffic operator.}
    \label{fig:framework}
\end{figure*}

In the following, we introduce a T2G prediction framework that allows a generic application to any intersection, providing traffic signal and LD data. The architecture is depicted in Figure~\ref{fig:framework}. The blocks (1) and (3) denote a supervised ML problem's processing and implementation steps. 

The raw data (i.e., LD and signal data from the traffic operator) functions as an input to the data pre-processing (Block (1)). The input signals are transformed into a structured format within this step, and undefined signal states are eliminated. Consequently, the quantities in the processed data set can be defined as follows: Let $s_i(k, c_{i,n})$ and $d_j(k, c_{i,n})$ be the signal state of a traffic signal $i \in \mathcal{S}$ and an LD $j \in \mathcal{D}$ at discrete time step $k$, respectively. Consequently, $s_i(k, c_{i,n})$ is defined as follows:
\begin{equation}
     s_i(k, c_{i,n}) = \left\{
        \begin{array}{ll}
            0, & \quad \text{if: $i$ is red } \\
            1, & \quad \text{else: $i$ is green.}
        \end{array}
    \right.
    \label{eq:def_1}
\end{equation}
Note that in (\ref{eq:def_1}) only the red and green signal phases are considered. Other common signal indications such as the start and end of a green phase (yellow and red-yellow, respectively~\cite{ref:riedel_2019}) are considered as $s_i(k, c_{i,n}) = 0$.  Analogously, we define the state $d_j(k, c_{i,n})$: 
\begin{equation}
     d_j(k, c_{i,n}) = \left\{
        \begin{array}{ll}
            0, & \quad \text{if: $j$ is not occupied} \\
            1, & \quad \text{else: $j$ is occupied}
        \end{array}
    \right.
    \label{eq:def_2}
\end{equation}

The final processed time series for all $i$ and $j$ are concatenated in the set $\textbf{R} = \{\{s_i(k, c_{i,n})\}_{i=1}^A, \{d_j(k, c_{i,n})\}_{j=1}^B\}_{n}$ which represents the non-aggregated data set (Figure~\ref{fig:framework}). $\textbf{R}$ serves as an input to Block (2), where the data set is aggregated and feature engineering is performed.

\subsection{Feature engineering}
\label{sec:feature_engineering}
\begin{figure}[!b]
    \centering
    \includegraphics[width=0.45\textwidth]{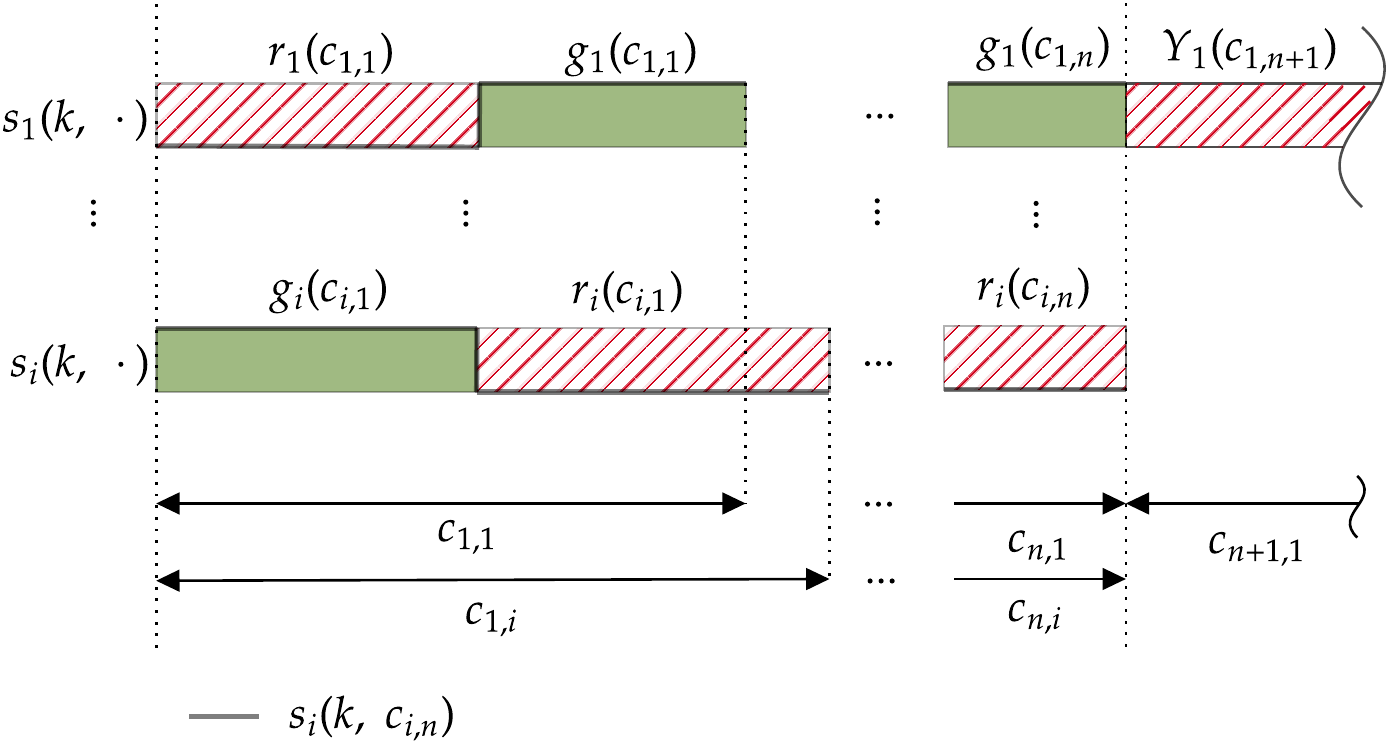}
    \caption{Feature computation of red and green time based on $s_i(k, c_{i,n})$. }
    \label{fig:signal_cycle}
\end{figure}

We perform data aggregation and feature engineering based on $\textbf{R}$. We aggregate the data by signal cycles. This approach is selected as (a) the prediction target T2G is an aggregated quantity by definition (a float value representing the duration of the next red phase) and (b) aggregated quantities are more easily accessible for traffic operators or other authorities compared to data streams with a resolution of, e.g., 1 sec. Note that this approach differs from previous works in~\cite{ref:genser_t2g_itsc, ref:genser2020enhancement}, where the non-aggregated data set $\textbf{R}$ is utilized without any further feature engineering. We first utilize the traffic signal data $s_i(k, c_{i,n})$ to compute the red $r_i(c_{i,n})$ and green time $g_i(c_{i,n})$ of a signal $i$ operating in cycle $c_{i,n}$. Consequently, the duration of the individual signal phases can be defined as follows: 

\begin{equation}
    r_i(c_{i,n}) = \sum_{k=1}^{K} \Big ( 1 -  s_i(k, c_{i,n}) \Big ),
    \label{eq:redtime}
\end{equation}

\begin{equation}
    g_i(c_{i,n}) = \sum_{k=1}^{K} s_i(k, c_{i,n}),
    \label{eq:greentime}
\end{equation}
Note that $K$ defines the discrete time step of the last sample of cycle $c_{i,n}$. Figure~\ref{fig:signal_cycle} depicts an example of the introduced quantities. The black pulse signals denote the raw traffic signal data $s_i(k, c_{i,n})$. The computation of the red and green time for signal $i=1$ by utilizing (\ref{eq:redtime}) and (\ref{eq:greentime}) give $r_1(c_{1,1})$ and $g_1(c_{1,1})$. The summation $r_1(c_{1,1}) + g_1(c_{1,1})$ results in the duration of cycle $c_{1,1}$. The derivation is performed for all traffic signals in $\mathcal{S}$ and serves as an input to forecast $Y_{i}(c_{i,n})$. Note that in Figure~\ref{fig:signal_cycle} no prediction is shown for $s_i(k, c_{i,n})$ as the last signal phase shows a red phase. Since we only predict the T2G, predictions of following green phases are not considered. Nevertheless, the framework would allow for such an application. 

Per definition, every $c_{i,n}$ starts with a red phase. To also utilize the temporal component of the set $\textbf{R}$, we compute the day, hour, minute, and second of a cycle's starting point as separate features denoted as $D$, $H$, $M$, and $S$, respectively. 

To enhance the prediction of the next red phase, data from LDs are of great importance as the detections transmitted to the signal control system are, in fact, key for determining future red- and green phases. Consequently, we utilize all signals $d_j(k, c_{i,n})~\forall j, k, i, n$ of detectors $j$ and compute a set of features to infer the current traffic state at the signalized intersection. Figure~\ref{fig:det_cycle} depicts an example of the utilized signals and visually supports the feature engineering in the following. First, we compute the traffic flow when traffic signal $i$ is red or green, respectively. To determine the traffic flow based on LD data, we assume that one signal peak corresponds to one detected vehicle. This is a reasonable assumption based on the time intervals used. Hence, the arrows in Figure~\ref{fig:det_cycle} indicate when vehicles pass a given LD, which corresponds to $d_j(k, c_{i,n})$ changing its state from 1 to 0. We denote these traffic flows during a red or green phase as $q_{i,\mathrm{R}}(c_{i,n})$ and $q_{i,\mathrm{G}}(c_{i,n})$. The quantities represent the summation of signal changes in the corresponding traffic signal phase defined with Iverson brackets (the function takes the value 1 if the statement is true and 0 otherwise) as follows: 
\begin{figure}[!b]
    \centering
    \includegraphics[width=0.45\textwidth]{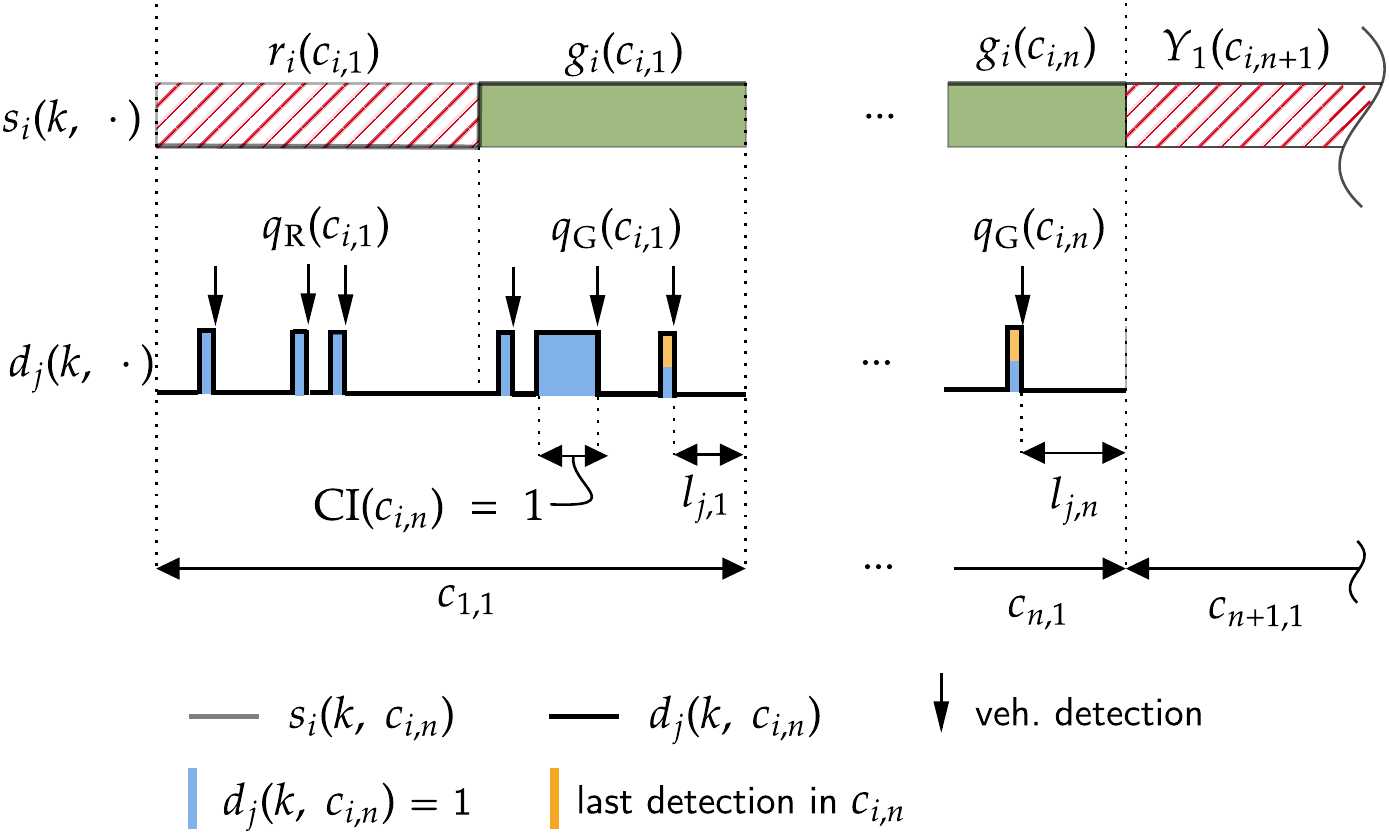}
    \caption{Feature computation based on traffic signal and detector data $s_i(k, c_{i,n})$ and $d_j(k, c_{i,n})$. }
    \label{fig:det_cycle}
\end{figure}

\begin{equation}
    q_{i,\mathrm{R}}(c_{i,n}) = \sum_{k_{\mathrm{R}} = 2}^{K_{\mathrm{R}}} \llbracket d_j(k_{\mathrm{R}} - 1, c_{i,n}) - d_j(k_{\mathrm{R}} , c_{i,n}) = 1 \rrbracket,
    \label{eq:q_red}
\end{equation}
and 
\begin{equation}
       q_{i,\mathrm{G}}(c_{i,n}) = \sum_{k_{\mathrm{G}} = 2}^{K_{\mathrm{G}}} \llbracket d_j(k_{\mathrm{G}} - 1, c_{i,n}) - d_j(k_{\mathrm{G}} , c_{i,n}) = 1 \rrbracket,
    \label{eq:q_green}
\end{equation}
where $K_{\mathrm{R}} = r_i(c_{i,n})$ and $K_{\mathrm{G}} = g_i(c_{i,n})$, i.e., the duration of the red and green phase, respectively. 

Next, we compute the occupancy of a detector $j$ during a cycle $c_{i,n}$. For that, we again utilize the corresponding signal $d_j(k, c_{i,n})$ and compute the summation of time steps the detector was occupied (indicated in Figure~\ref{fig:det_cycle} by the blue areas). The summation is then normalized by the cycle length of $c_{i,n}$ which is computed by the summation of cycles' red and green time: 

\begin{equation}
    o_j(c_{i,n}) = \frac{\sum_{k=1}^{K} d_j(k, c_{i,n}) }{r_i(c_{i,n}) + g_i(c_{i,n})}
    \label{eq:occ}    
\end{equation}

The occupancy is defined within the interval $[0,1]$; if a detector is not occupied within a signal cycle, the occupancy is 0; if a detector is fully occupied throughout a cycle, $o_j(c_{i,n}) = 1$. 

Further, we want to gain information about the last detection of an LD in a signal cycle. This feature allows for inferring information about the current traffic demand at a signal. Especially for public transportation vehicles, this feature can be utilized as a proxy to provide information about the next arrival. For example, suppose a detector that only gives detection information about a bus or tram has been activated in the last cycle. In that case, the likelihood that no detections occur in the next cycle might increase, and consequently, a longer red phase might be expected. The orange areas highlight the last detection of a detector in Figure~\ref{fig:det_cycle}. We compute the time duration from the last detection to the end of a cycle $l_{j}$ as follows: 

\begin{equation}
    l_{j}(c_{i,n}) = \Big (  r_i(c_{i,n}) + g_i(c_{i,n})  \Big) - v(d_j(k,c_{i,n})),
    \label{eq:last_det}
\end{equation}
where the function $v(\cdot)$ computes the time stamp of the last detection in cycle $c_{i,n}$ of detector $j$ based on the signal $d_j(k, c_{i,n})$.

As last feature inputs, we compute a queue and congestion indicator denoted as $\mathrm{QI}_i(c_{i,n})$ and $\mathrm{CI}_i(c_{i,n})$, when traffic light $i$ is red or green, respectively. The quantities' definition is based on a threshold of the time duration of a single detector activation; i.e., if $d_j(k, c_{i,n})$ shows an activation lasting longer than a threshold parameter $p$ in seconds during a cycle $c_{i,n}$, dependent on the operated signal phase, $\mathrm{QI}_i(c_{i,n})$ or $\mathrm{CI}_i(c_{i,n})$ is set to 1. Formally, this can be denoted as: 
\begin{equation}
    \mathrm{QI}_i(c_{i,n}) = \llbracket u(d_j(k, c_{i,n})) > p \wedge s_i(k, c_{i,n}) = 0  \rrbracket,
    \label{eq:QI}
\end{equation}
and 
\begin{equation}
    \mathrm{CI}_i(c_{i,n}) =  \llbracket u(d_j(k, c_{i,n})) > p \wedge s_i(k, c_{i,n}) = 1 \rrbracket,
    \label{eq:CI}
\end{equation}
where function $u$ determines the longest detection during a cycle and computes the corresponding time duration in seconds. The returned set of values from $u$ is then thresholded with $p$ and conditional on the state of $s_i(k, c_{i,n})$; if the long occupation represents a queue during red light or congestion during a green light. We utilize the queue and congestion indicators to determine if a single vehicle or multiple vehicles (with small headway) occupy a detector longer than $p$. The latter does represent a traffic state where queues/congestion is likely. However, theoretically, the occupation  larger than $p$ caused by a single vehicle can also be caused by a random phenomenon (e.g., a taxi loading/unloading passengers). Therefore, this does not necessarily represent the same traffic state as the occupation by multiple vehicles. Nevertheless, this behavior can lead to queues/congestion, so we treat these two cases identically. 

Finally, we can derive the target variable $Y_i(c_{i,n})$. We predict the next T2G based on an input sample from the current cycle. As the T2G target value in the data constitutes the red time of the next cycle $r_i(c_{i,n+1})$, the target feature is simply denoted as: 
\begin{equation}
    Y_i(c_{i,n}) = r_i(c_{i,n+1})
\end{equation}

The data set combined in $\textbf{X}_n$ contains the red times $r_i(c_{i,n})$, green times $g_i(c_{i,n})$, traffic flow during a red and green phase $q_{i,\mathrm{R}}(c_{i,n})$, $q_{i,\mathrm{G}}(c_{i,n})$, the LD occupancy $o_j(c_{i,n})$, time since last detections during a cycle $l_{j}(c_{i,n})$, and the queue and congestion indicators $\mathrm{QI}_i(c_{i,n})$, $\mathrm{CI}_i(c_{i,n})$ for all traffic lights $i$. Finally, the T2G values $Y_i(c_{i,n})$, serving as targets for the regression problem, are added to $\textbf{X}_n$ and utilized to implement the supervised learning problem with a set of machine learning models for tackling $\hat{Y}(c_{i,n}) = f(\textbf{X}_n)$.

\subsection{Naive baseline and linear model}
\label{sec:baselines}
The naive model is introduced as a first baseline model, where the prediction of the next T2G is simply set to the last observed red time. Formally, this can be denoted as follows: 

\begin{equation}
    \begin{split}
         \hat{Y}_{i}(c_{i,n}) = r_i(c_{i,n}).
    \end{split}
    \label{eq:no_change}
\end{equation}
Such a simple forecasting approach is utilized in various research domains and also transportation~\cite{ref:tan_naive_traffic_flow, ref:smith_models_flow_forecast} for performance comparison of more robust forecasting models. As stated by~\cite{ref:mclaughlin1983forecasting}, naive models should not be treated as forecasting models but rather as a benchmark to disqualify proposed prediction models that perform worse on a problem than a naive model. 

Secondly, we introduce the LR, which can be defined as follows:
\begin{equation}
    \begin{split}
         \hat{Y}_{i}(c_{i,n}) = \beta_{i,0} + \beta_{i,1} x_1(c_{i,n}) + \beta_{i,2} x_2(c_{i,n}) + ... \\
          + \beta_{i,p} x_p(c_{i,n}) + E_i(c_{i,n}), \ \forall o = 1,..,T,
    \end{split}
    \label{eq:LR}
\end{equation}
where $\hat{Y}_{i}(c_{i,n})$ is the T2G (response variable) for signal $i$ and cycle $c_{i,n}$, $ \beta_{i,0}$ represents the intercept term and $\beta_{i,1} $ to $\beta_{i,p}$ are the regression coefficients for the $p$ predictors $x_1(c_{i,n})$ to $x_p(c_{i,n})$ (i.e., the members of the LD and signal feature set $\textbf{X}_n$ described above), respectively. The error term is denoted by $E_i(c_{i,n})$ and follows a Gaussian distribution (i.e., $E_i(c_{i,n}) \sim \mathcal{N}(0, \sigma_{E_i(c_{i,n})}$); $T$ denotes the prediction horizon. The solution for $\hat{Y}_i(c_{i,n})$ is found by applying the Ordinary Least Square (OLS) method. The fitted model can be used to determine a prediction of the T2G for a given traffic signal by obtaining the conditional expected value of the response. To obtain the LR model, the implementation from \textit{scikit-learn}~\cite{ref:scikit-learn} is utilized. Research that similarly introduces LR models within this context can be found in e.g.,~\cite{ref:LR1}.

\subsection{Supervised learning model candidates}
\label{sec:mach_learn_models}
We introduce the RF, a supervised learning technique based on ensemble learning utilizing decision trees based on the work from~\cite{ref:breiman2001random}. RFs are applied in various research domains for classification and regression tasks. In this work, we utilize the RF implementation from \textit{scikit-learn}~\cite{ref:scikit-learn} to predict the T2G by solving a supervised regression problem. The implemented procedure is described with the pseudo-code in Algorithm~\ref{ag:1}. First, samples are randomly selected with replacements from the training data set to create a bootstrap sample, which is a member of $\mathcal{B}$; i.e., one sample can be selected more than once. Next, for all bootstrap samples, a decision tree is fit. This procedure results in a collection of decision trees that are denoted as an RF. Before the average prediction error from all decision trees is calculated, the Out-of-Bag (OOB) data set is collected. None of the data samples belonging to the OOB are selected during the computation of the randomized process in Step 2. Finally, the prediction error on the OOB data is calculated for all trees and averaged. 

\begin{algorithm}[!t]
\caption{RF pseudo code}\label{euclid}
\label{ag:1}
\begin{algorithmic}[1]
\Procedure{doRandomForest}{}
\State 
     \begin{varwidth}[t]{\linewidth}
        $\mathcal{B} \gets$ Bootstrap samples from the  \par
        training data set randomly
        \end{varwidth}

\State 
     \begin{varwidth}[t]{\linewidth}
        $\mathcal{T} \gets$ Grow and fit a decision tree $\forall s \in \mathcal{B}$, \par
        where $s$ is a bootstrap sample from $\mathcal{B}$
        \end{varwidth}

\State $\mathcal{O} \gets \text{Exclude out-of-bag data}$

\State 
     \begin{varwidth}[t]{\linewidth}
     $\mathcal{C} \gets$ Calculate the average prediction
     $\forall t \in \mathcal{T}$, \par where $t$ is a decision tree for one bootstrap sample 
     \end{varwidth}
\State $\mathcal{P} \gets \text{Calculate average prediction error using~} \mathcal{O}$
\EndProcedure
\end{algorithmic}
\end{algorithm}

For a detailed mathematical background on RF and corresponding theorems and proofs, the interested reader is referred to~\cite{ref:breiman2001random}.

The last ML model incorporated into the T2G framework is an LSTM network, a particular type of RNN. To address the drawbacks of standard memory-less RNNs (vanishing gradient or exploding), extensions regarding the network architecture with a memory block were proposed by~\cite{ref:hochreiter_LSTM}. Along with other neural network designs, an LSTM is constructed with an input, hidden, and output layer. In addition, the hidden layer is designed with a memory block containing memory cells. The state of these cells is influenced by memorizing the temporal state and gating units that control the information flow in one memory cell. 
In addition, input and output gates are implemented to control the input and output activation, respectively. When the information state of a memory cell is outdated, a forget gate allows an automatic reset to forget information that loses importance while evolving over time~\cite{ref:Ma_LSTM}. The model formulation is denoted with an input $x = (x_{i,1}, x_{i,2},..., x_{i,K})$ and the output $y = (y_{i,1}, y_{i,2}, ...y_{i,K})$. $y_k$ is the predicted response, and $K$ is the prediction horizon. To compute the model response for the next time step, the following equations are introduced. For simplicity, note that the index for signal $i$ and cycle $c_{i,n}$ are omitted; also the variables introduced here are internal model variables and should not be mistaken with the feature variables above:
\begin{equation}
    i_k = \mathrm{sig} \Big ( W_{ix}x_k + W_{im}m_{k-1} + W_{ic}c_{k-1} + b_i \Big ),
    \label{eq:lstm_1}
    \end{equation}
    \begin{equation}
    f_k = \mathrm{sig} \Big ( W_{fx}x_k + W_{fm}m_{k-1} + W_{fc}c_{k-1} + b_f \Big ),
    \label{eq:lstm_2}
    \end{equation}
    \begin{equation}
    c_k = f_k \odot c_{k-1} + i_k \odot g \Big(W_{cx}x_k + W_{cm}m_{k-1} + b_c \Big ),
    \label{eq:lstm_3}
    \end{equation}
    \begin{equation}
    o_k = \mathrm{sig} \Big(W_{ox}x_k + W_{om}m_{k-1} + W_{oc}c_{k} + b_o \Big ),
    \label{eq:lstm_4}
    \end{equation}
    \begin{equation}
    m_k = o_k \odot h(c_{k}),
    \label{eq:lstm_5t}
    \end{equation}
    \begin{equation}
    y_k = W_{ym}m_k + b_{y},
\label{eq:lstm_6}
\end{equation}
where $i_k$, $f_k$, $c_k$, $o_k$ and $m_k$ are the states of the input gate, forget gate, cell state, output gate and memory gate, respectively. The variables $W$ and $b$ denote the weight matrices and bias vectors, respectively, and are utilized to connect input, hidden, and output layers. Note that
$\mathrm{sig}(\cdot)$ defines the logistic function (i.e., sigmoid function); $g(\cdot)$ and $h(\cdot)$ denoted activation functions, respectively, where commonly $\tanh$ is utilized~\cite{ref:Ma_LSTM}. The work in \cite{ref:hochreiter_LSTM} introduces similar mathematical descriptions of LSTM networks. The implementation in our framework is performed with TensorFlow~\cite{ref:tensorflow} and Keras~\cite{ref:keras}. 

\subsection{Hyperparameter tuning}
\label{sec:hyperparameter_tuning}
Hyperparameter tuning (block (3) in Figure~\ref{fig:framework}) is an essential step in a machine learning pipeline to improve the model accuracy. Therefore, we utilize two libraries to finalize the RF and LSTM models in this work. To train the hyperparameters of the RF models, we utilize the open-source library Hyperopt~\cite{ref:bergstra_hyperopt}. The framework allows defining a hyperparameter search space. Then one of the implemented optimization algorithms is utilized to sample values from the pre-specified distributions and evaluate the model for several trail runs. Every trail model is then evaluated with the specified loss function during k-fold cross-validation. In this work, we choose to optimize the following hyperparameters of the RF models: the number of estimators, max depth, min samples split, and min weight fraction leaf. As an optimization algorithm, the adaptive Tree-structured Parzen Estimator (TPE) is utilized~\cite{ref:bergstra_hyperopt}.

For the hyperparameter tuning of the LSTM neural network models, we utilize KerasTuner~\cite{ref:omalley2019kerastuner}, built in the deep learning API Keras. KerasTuner is an optimization framework for the tuning of hyperparameters with state-of-the-art algorithms. The framework allows to define ranges for non-conditional and conditional hyperparameters, from which values are sampled during a trial run. To find the best model performance for all LSTM neural networks, we optimize the following hyperparameters: number of units for the LSTM input and output layer, number of hidden LSTM layers with the corresponding number of units, dropout rate, and number of units for dense layer. Note that we fix the activation function of the dense layer and utilize the ReLU function~\cite{ref:LeCun_DeepLearning}. As an optimization algorithm, the Hyperband Tuner proposed by~\cite{ref:hyperband} is applied. Note that in both tuning procedures the MAE and Mean Squared Error (MSE) are utilized as a loss function, and the models with the best results are selected. 

\subsection{Performance metrics}
\label{sec:performance_metrics}
We evaluate the models on the test data set $\textbf{X}_{n,\mathrm{test}}$ with performance metrics in block (4). First, the Mean Absolute Error (MAE) and the Root Mean Square Error (RMSE) are utilized. The performance metrics are introduced by (\ref{eq:per_1}) and (\ref{eq:per_3}):
\begin{equation}
    {\rm MAE} = \frac{1}{K_{\mathrm{test}}}\sum_{k=1}^{K_{\mathrm{test}}} \Big |\hat{Y}_{i}(c_{i,n+k}) - 
 Y_{i}(c_{i,n+k}) \Big |,
    \label{eq:per_1}
\end{equation}

\begin{equation}
    {\rm RMSE} = \sqrt{\frac{1}{K_{\mathrm{test}}}\sum_{k=1}^{K_{\mathrm{test}}} \Big |\hat{Y}_{i}(c_{i,n+k}) - Y_{i}(c_{i,n+k}) \Big |^2}.
    \label{eq:per_3}
\end{equation}
$\hat{Y}_{i}(c_{i,n+k})$ again represents the predicted T2G for a future cycle $c_{i,n+k}$ of signal $i$. $Y_{i}(c_{i,n+k})$ is the T2G from the test data set. $k$ is here utilized to sum the errors over all samples from the test data set, i.e., $K_{\mathrm{test}}$. 
Besides the evaluation concerning the MAE and RMSE, we introduce two additional and strict error metrics. As we want to evaluate if the prediction meets the requirements of practical applications (e.g., speed-advisory systems), we introduce the Exact Hit (EH) and the Near-Misses (NM) ratio as follows: 

\begin{equation}
    \mathrm{EH} = \Bigg ( \frac{\sum\limits_{k=1}^{K_{\mathrm{test}}} \Big \llbracket \Big | \hat{Y}_{i}(c_{i,n+k}) - Y_{i}(c_{i,n+k}) \Big | = 0 \Big \rrbracket }{K_{\mathrm{test}}}\Bigg) \cdot 100,
    \label{eq:EH}
\end{equation}

\begin{equation}
    \mathrm{NM} = \Bigg ( \frac{\sum\limits_{k=1}^{K_{\mathrm{test}}} \Big \llbracket \Big | \hat{Y}_{i}(c_{i,n+k}) - Y_{i}(c_{i,n+k}) \Big | \leq 2 \Big \rrbracket }{K_{\mathrm{test}}}\Bigg) \cdot 100.
    \label{eq:NM}
\end{equation}
An EH is defined when the prediction model forecasts the T2G with an error of 0 seconds and an NM when the error is lower or equal to two seconds. The threshold value is chosen based on studies such as e.g.,~\cite{ref:brunner, ref:makridis}, that find the response time for action of Connected Automated Vehicles (CAV) is ranging from one to two seconds (and higher under certain circumstances). Consequently, predictions that are classified as EH or NM can serve as an input to, e.g., motion planning algorithms without forcing a vehicle to stop at an intersection. Note that the T2G values of the test data set are given as integer values; therefore, we also round the prediction values to integers. 

\section{NUMERICAL EXPERIMENTS}
\label{sec:num_exp}
\begin{figure}[t]
    \centering
    \includegraphics[width=0.5\textwidth]{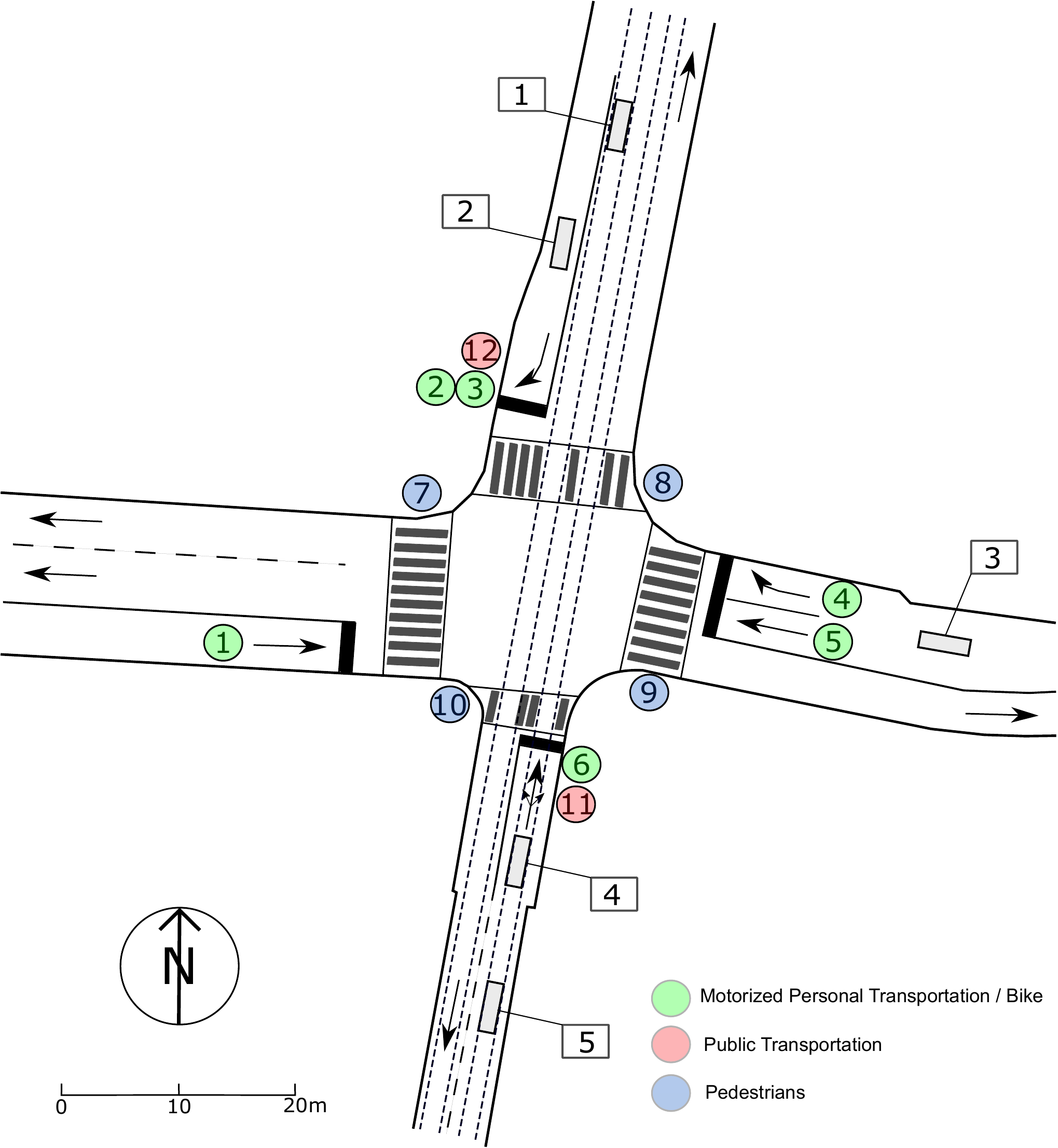}
    \caption{Test intersection in the city center of Zurich, Switzerland. }
    \label{fig:intersection}
\end{figure}

The numerical experiment is based on a historical data set from an intersection in the city center of Zurich, Switzerland. The intersection depicted in Figure~\ref{fig:intersection} is regulated by a fully-actuated signal control system with 12 traffic signals (indicated with circled numbers). Signals 1, 2, 4, 5, and 6 control vehicular traffic streams. Traffic signal 3 is implemented for cyclists who can travel straight ahead and is co-regulated with signal 2. Signals 7 -- 10 regulate pedestrian flows. From north to south and vice versa, tram lines are potentially prioritized by the signal control. Tram tracks are indicated with dashed lines and overlap with car lanes southbound of the intersection. The corresponding signals 11 and 12 are specifically implemented for trams. These signals only operate in a green phase when a public transportation vehicle arrives at the intersection. The case study does not focus on predicting the T2G for these signals, i.e., only results for signal 1 -- 10 are presented. A discussion with numerical insights for signals 11 and 12 is then presented in Section~\ref{sec:diss}.

Figure~\ref{fig:intersection} also depicts the five associated LDs (indicated as rectangles with the corresponding numbering at the intersection approaches). Note that no separate detector data is implemented for traffic signals 1 and 3. Hence, no information on arriving vehicles/cyclists is available. 

The data set contains event-based data from every device installed at the test site (i.e., LD and signal data). With a resolution of 1 second, a telegram is generated every time a device changes its state. A telegram contains the timestamp, device id, and the new state. In other words, we know when an LD is activated/not activated, or a signal operates in a red or green phase. Besides, LD or signal failures can be detected.

A raw data set of 2 months of consecutive telegram data from January and February 2019 is available. Note that the time axis of the data set is unevenly spaced as telegrams are only reported when an event occurs. 
In the following analysis, we take data from all weekdays from 7:00 to 22:00 hours into account as public transportation operates on a regular schedule within this time frame. 

\subsection{Data aggregation and descriptive analysis}
\label{sec:desc}
First, we process the telegram data set for 2 months to eliminate telegrams containing devices or values not defined for the case study intersection. Also, the raw data set contains clock telegrams (the device state is reported every full minute, regardless of a state change) that are not relevant for this study. Finally, the data is processed to a tabular format, required for machine learning purposes; i.e., we compute the data set $\textbf{R}$ containing all $s_i(k, c_{i,n})$ and $d_j(k, c_{i,n})$. 

Afterward, we proceed to aggregate the data set to cycles. As cycles diverge for every traffic signal $i$, we create a separate data set for every traffic signal. Every feature in such a data set is grouped by the corresponding cycles of $c_{i,n}$. This approach allows determining the features of all remaining traffic signals and LDs during the cycles of $i$. Consequently, we determine 10 data sets with the proposed feature set: red time, green time, traffic flow during red and green, occupancy, last detection of an LD, and the queue and congestion indicator. These features are concatenated to the vector $\textbf{X}_n$ and serve as an input to the model training/testing procedure. Last, we determine the T2G $Y_i({c_{i,n}})$ which serve as the target value for the regression problem. 

\begin{figure*}[!t]
   \centering
    \includegraphics[width=1\textwidth]{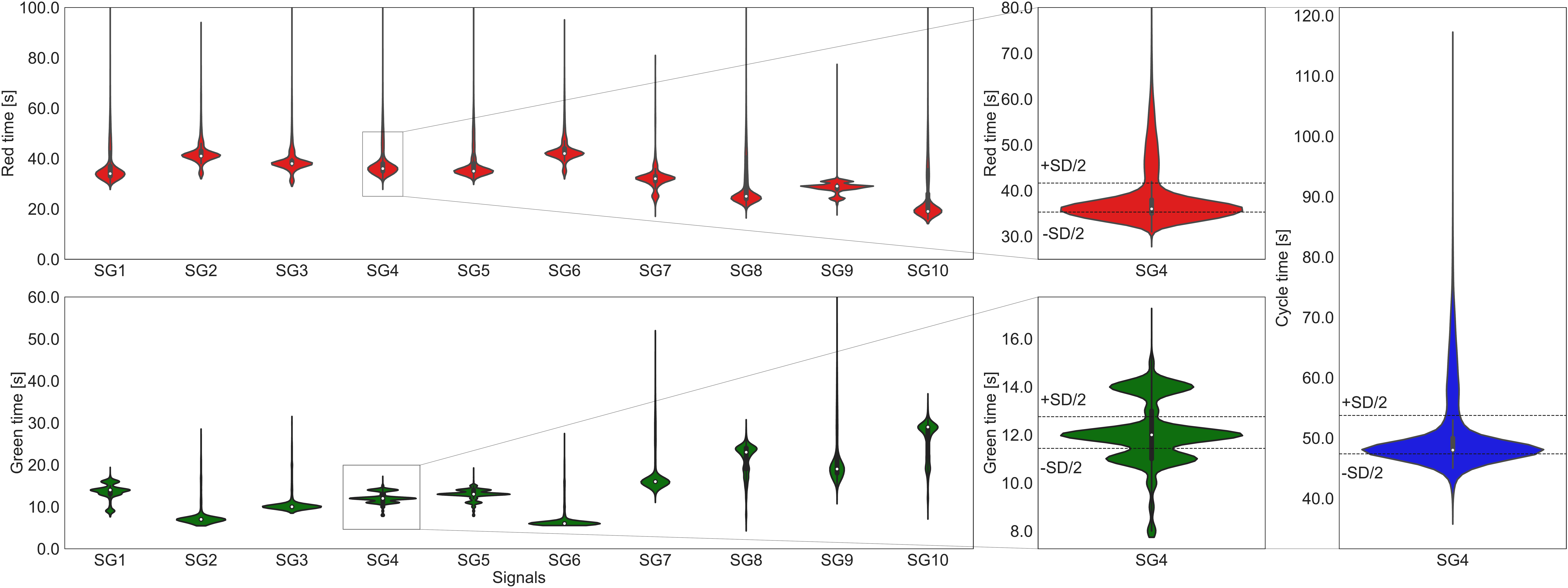}
  \caption{Distributions of red and green times for all signals 1--10. Additionally, the distributions for red, green, and cycle time of signal 4 are shown with corresponding standard deviation thresholds +SD/2 and -SD/2.}\label{fig:violines}
\end{figure*}

In Figure~\ref{fig:violines}, feature distributions for the red and green times of all traffic signals are presented.  

The violin plots present the red and green time feature distributions for all considered traffic signals. The data distributions highlight that the signal control system is fully-actuated. For example, the red time of signal 4 (shown separately in Figure~\ref{fig:violines}) operates with an average red time of 38.46 seconds. Nevertheless, the minimum and maximum values in the data set show red times of 29 sec and 104 sec, respectively. Note that signal 4's red time distribution shows a long tail due to red times higher than 70 seconds. Nevertheless, these samples are not outliers, as the maximum allowed red time for this signal control strategy is fixed to 180 sec. Hence, the prediction models must be capable of learning and predicting this behavior. The threshold values +SD/2 (41.63 sec) and -SD/2 (35.30 sec) highlight the range around the mean value of 38.46 sec. 28207 from 53031 available samples occur within the specified range. Consequently, 24824 samples of signal 4's red time are outside +SD/2 or -SD/2 and not close to the mean value. This underlines the full actuation of the system and that not just a few samples show a high variance. Similar characteristics can be shown for the green and cycle time distribution of signal 4. The range around the mean green time of 12.01 sec are computed with 12.75 and 11.44 sec. From 53031 samples, 24306 samples are within, and 28725 samples are outside the range between +SD/2 and -SD/2. For the cycle time with a mean value of 50.56 sec, 33323 samples are close to the mean and 19708 samples are outside the range. The descriptive analysis underlines that (a) the system is fully-actuated and (b) a substantial amount of data is different from the mean. Note that signal 4 is presented here as it regulates a traffic stream conflicting with public transportation (controlled by signal 12). Hence, for both traffic lights, the signalization is influenced by the arrival of trams. Additionally, LD 3, upstream from signal 4, might be utilized for green time extensions when high traffic demand is present. 

A complete compilation of the computed descriptive statistics of the data sets for all traffic signals is given in Table~\ref{tab:desc_stats}.
\begin{table*}[!t]
\centering
\setlength{\tabcolsep}{4pt}
\caption{Descriptive statistics of input features for traffic signals $i=[1,10]$ represented as mean values. The dash indicates that no detector is available and features are not computed. The values within parentheses indicate the ($\min$, $\max$) of the corresponding feature.}

\begin{tabular}{L{0.5cm}L{0.8cm}R{0.8cm}L{1.1cm}R{0.7cm}L{1.0cm}R{0.5cm}L{0.7cm}R{0.5cm}L{0.7cm}R{0.5cm}L{0.7cm}R{0.8cm}L{1.4cm}R{0.5cm}L{0.7cm}R{0.5cm}L{0.7cm}}
\toprule
$i$  & \# $c_{i,n}$ [-] & \multicolumn{2}{c}{$\Bar{r}_i$} [s] & \multicolumn{2}{c}{ $\Bar{g}_i$} [s] & \multicolumn{2}{c}{$\Bar{q}_{i,\mathrm{R}}$} [veh] & \multicolumn{2}{c}{$\Bar{q}_{i,\mathrm{G}}$} [veh] & \multicolumn{2}{c}{$\Bar{o}_j$} [-] & \multicolumn{2}{c}{$\Bar{l}_{\mathrm{det},j}$} [s] & \multicolumn{2}{c}{ $\Bar{\mathrm{CI}}_i$} [-] & \multicolumn{2}{c}{$\Bar{\mathrm{QI}}_i$} [-] \\
\midrule
1      & 64807       & 36.64       & (29, 102)      & 13.57        & (8, 19)         & -            & -             & -             & -              & -         & -           & -          & -               & -         & -          & -         & -          \\
2      & 67242       & 42.38       & (33, 102)      & 7.62         & (7, 28)         & 0.73         & (0, 8)        & 0.17          & (0, 6)         & 0.06      & (0, 1)      & 107.53     & (0, 2115)       & 0.07      & (0, 1)      & 0.03      & (0, 1)     \\
3      & 67916       & 39.22       & (30, 100)      & 10.79        & (9, 31)         & 0.68         & (0, 7)        & 0.22          & (0, 6)         & 0.06      & (0, 1)      & 107.39     & (0, 2115)       & 0.06      & (0, 1)     & 0.06      & (0, 1)     \\
4      & 53031       & 38.46       & (29, 104)      & 12.01        & (8, 17)         & 1.70         & (0, 9)        & 0.61          & (0, 6)         & 0.07      & (0, 1)      & 63.38      & (0, 4014)       & 0.02      & (0, 1)     & 0.12      & (0, 1)     \\
5      & 50937       & 38.00       & (31, 105)      & 12.89        & (8, 19)         & 1.71         & (0, 8)        & 0.62          & (0, 7)         & 0.07      & (0, 1)      & 63.57      & (0, 4010)       & 0.06      & (0, 1)     & 0.11      & (0, 1)     \\
6      & 63712       & 43.40       & (33, 94)       & 6.70         & (6, 27)         & 0.18         & (0, 2)        & 0.02          & (0, 1)         & 0.04      & (0, 1)      & 196.87     & (0, 4815)       & 0.06      & (0, 1)      & 0.07      & (0, 1)      \\
7      & 67909       & 32.28       & (18, 80)       & 17.74        & (12, 56)        & -            & -             & -             & -              & -         & -           & -          & -               & -         & -          & -         & -          \\
8      & 67902       & 28.79       & (18, 99)       & 21.23        & (5, 32)         & -            & -             & -             & -              & -         & -           & -          & -               & -         & -          & -         & -          \\
9      & 67913       & 28.48       & (18, 79)       & 21.53        & (12, 87)        & -            & -             & -             & -              & -         & -           & -          & -               & -         & -          & -         & -          \\
10     & 67902       & 24.21       & (16, 101)      & 25.80        & (8, 38)         & -            & -             & -             & -              & -         & -           & -          & -               & -         & -          & -         & -          \\
\bottomrule
\end{tabular}

\label{tab:desc_stats}
\end{table*}
The red and green times of signals 1 -- 10 all show maximum values greater or equal than 80 sec. Contrary, the maximum values of green times are below 56 seconds, except for signal 9 (87 sec.). The high green time of signal 9 occurs because this pedestrian traffic signal can remain green together with signals 11/12. 

\subsection{T2G prediction results}
\label{sec:results}
This section applies our set of models to the training and testing procedure. We assess the model qualities by utilizing the processed data set and split it into 70\% train and 30\% test data, respectively. First, the naive baseline model is applied to the train data sets of traffic lights $i=[1,10]$. As discussed, the naive baseline model is a benchmark to assess the ML models applied in the following. As the model utilizes the last red time of a signal cycle, no training or hyperparameter training procedure is performed. 
\begin{table}[tb]
\centering
\caption{Model performance on the test data set with $\rm MAE$, $\rm RMSE$, $\mathrm{EH}$, and $\mathrm{NM}$ for the naive baseline model.}

\begin{tabular}{lrrrr}
\toprule
       & \multicolumn{4}{c}{Naive Baseline}                     \\
$i$ & $\mathrm{MAE}$ {[}s{]} & $\mathrm{RMSE}$ {[}s{]} & $\mathrm{EH}$ {[}\%{]} & $\mathrm{NM}$ {[}\%{]} \\
\midrule
1      & 5.22        & 9.20         & 35.49       & 58.43               \\
2      & 4.59        & 7.89         & 36.53       & 55.93              \\
3      & 4.59        & 7.93         & 37.16       & 56.07               \\
4      & 5.48        & 9.29         & 33.37       & 56.30               \\
5      & 4.30        & 8.37         & 46.42       & 67.41               \\
6      & 4.87        & 8.24         & 33.73       & 54.92               \\
7      & 4.21        & 7.20         & 36.06       & 55.59               \\
8      & 7.32        & 11.75        & 33.33       & 48.51               \\
9      & 1.54        & 2.78         & 54.68       & 74.26               \\
10     & 8.42        & 13.48        & 36.81       & 49.35               \\
\bottomrule
\end{tabular}
\label{tab:res_naive}
\end{table}

\begin{figure}
    \centering
  \subfloat[]{%
       \includegraphics[width=0.8\linewidth]{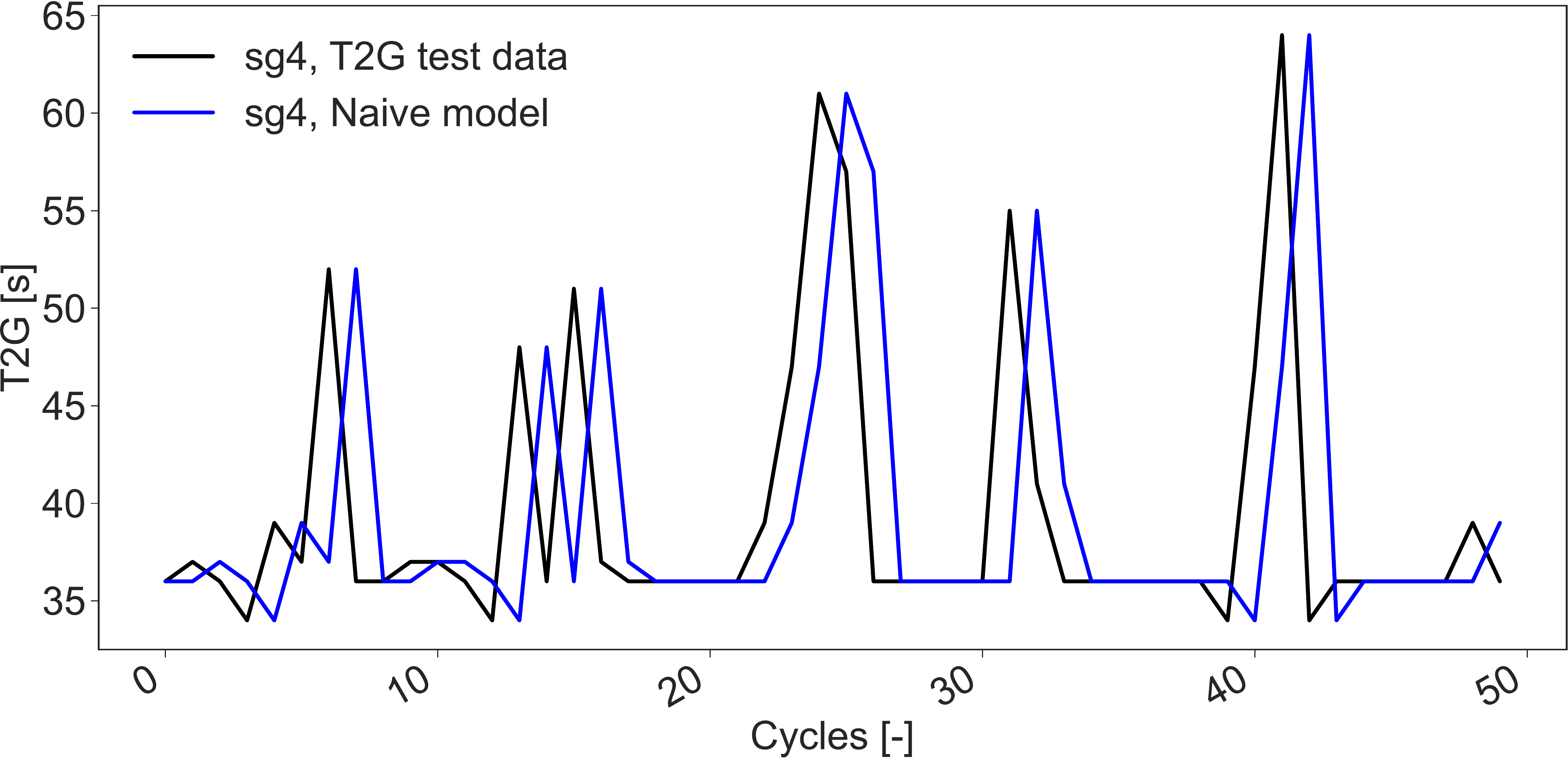}} \\
  \subfloat[]{%
        \includegraphics[width=0.8\linewidth]{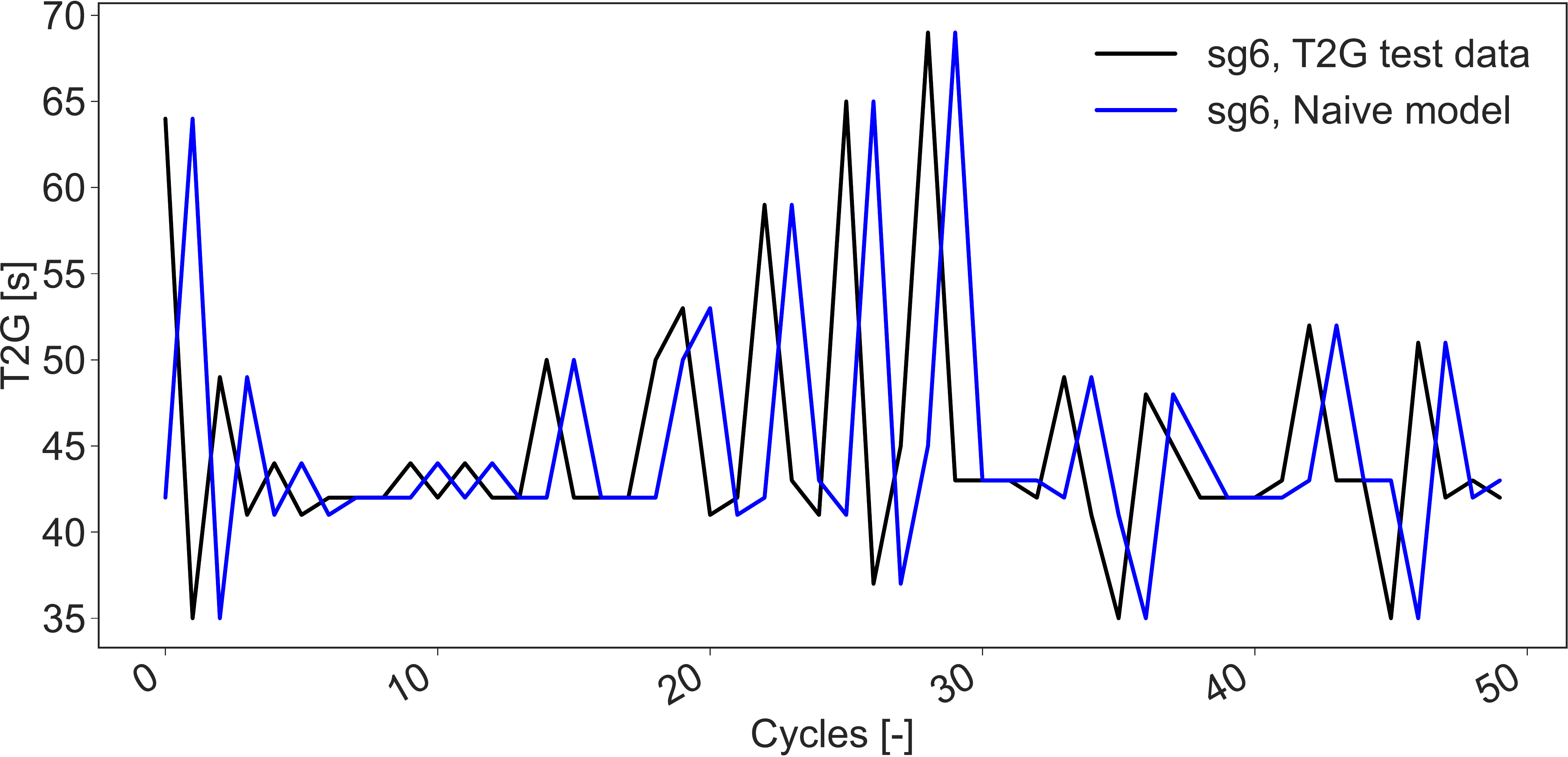}}
        
  \caption{Prediction results of the naive baseline model. Results assessed on the T2G test data set for (a) traffic signal 4 and (b) traffic signal 6.}
  \label{fig:result_comp} 
\end{figure}

For traffic lights 1 -- 10, the results of the naive baseline show MAE and RMSE errors from 1.54/2.78 sec ($i=9$) to 8.42/13.48 sec ($i=10$). Also, the EH ratio is below 37.16\%  for all traffic lights except signals 5 and 9 (ratio of 46.42\% and 54.68\%, respectively). The NM ratios are below 60\%, except for traffic signals 5 and 9, where 67.41\% and 74.26\% are computed, respectively. See Table~\ref{tab:res_naive} for all results. The highest NM ratio is computed for traffic light 9 which also shows the highest EH ratio of 54.68\%. The difference in performance can be explained by the variation of the T2G values: If the signal control assigns the identical green phase multiple times throughout a certain time frame (such a system behavior can correspond to a standard program; i.e., no high traffic demand detected or arriving public transportation), the naive model predicts an exact hit with an error of 0.00 sec. Contrary, the absence of a standard control program or high variations in the T2G lead to an obvious worse performance of the naive model. 
A subset of the prediction results (50 cycles) assessed with the corresponding test data subset are depicted in Figure~\ref{fig:result_comp}. Results show (a) low variability of the T2G for traffic light 4 and (b) high variability in the T2G for traffic signal 6. Note that the prediction values of the T2G derived by utilizing the naive model constitute a shift by one cycle. The presented time frames of prediction results are utilized throughout this section to ensure comparability.

In particular, Figure~\ref{fig:result_comp} (a) shows time frames where the T2G stays constant throughout multiple cycles. In these cases, the naive model performs with high accuracy, whereas in Figure~\ref{fig:result_comp} (b), high variations with fluctuations of the T2G between 37 sec and 68 sec are present. In these cases, the time lag of one cycle produces high error rates. 

We now present the results of the LR models for all traffic lights, respectively, which serve as a supervised learning baseline in this study. Before all models are trained, the Recursive Feature Elimination (RFE) method is applied to the traffic signal data sets. The method assigns a weight to an input feature which functions as a proxy for feature importance. The features with minor importance are then eliminated from the data set. In our case, the feature space is reduced from 70 features to 35 features. For example, two traffic signals that regulate non-conflicting pedestrian flows within the same signal stage show a high correlation, and RFE will eliminate one of the features. 

\begin{figure}
    \centering
  \subfloat[]{%
       \includegraphics[width=0.8\linewidth]{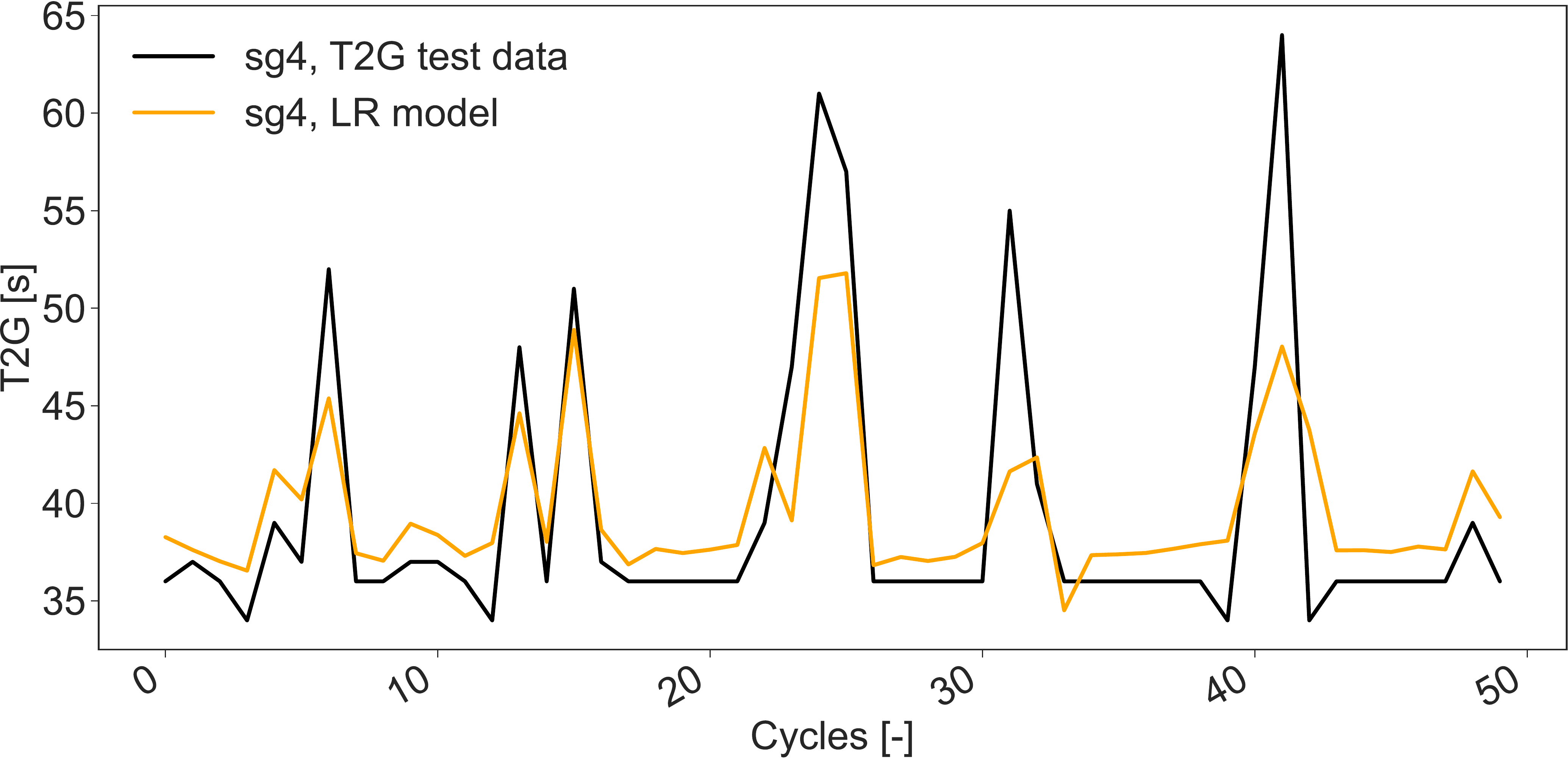}} \\
  \subfloat[]{%
        \includegraphics[width=0.8\linewidth]{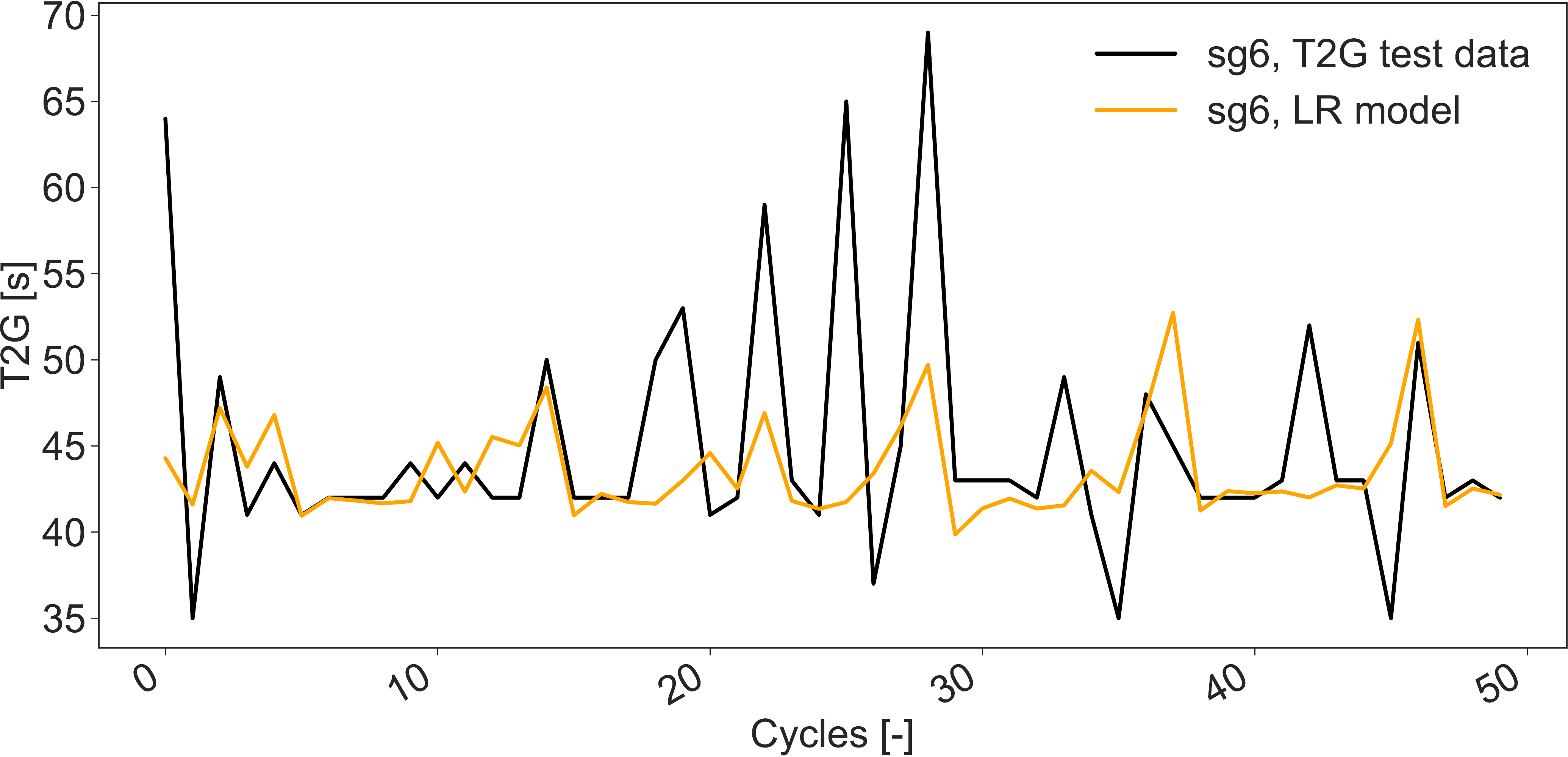}}
        
  \caption{Prediction results of the LR model. Results assessed on the T2G test data set for (a) traffic signal 4 and (b) traffic signal 6.}
  \label{fig:result_comp2} 
\end{figure}

All LR models are derived by the application of the OLS method and the prediction performance of the T2G samples of the test data set are assessed with the introduced performance metrics. The performance results are compiled in Table~\ref{tab:res_lr} and prediction results for the same test data subset are presented in Figure~\ref{fig:result_comp2}.

\begin{table*}[!t]
\centering
\caption{Model performance on the test data set with $\rm MAE$, $\rm RMSE$, $\mathrm{EH}$, and $\mathrm{NM}$ for the LR model. Also, the improvemenst over the naive baseline are presented with the given performance metrics.}
\begin{tabular}{rrrrr|rrrr}
\toprule
       & \multicolumn{4}{c}{LR model}                           & \multicolumn{4}{c}{Improvement over naive baseline}    \\
       \midrule
Signal & MAE {[}s{]} & RMSE {[}s{]} & EH {[}\%{]} & NM {[}\%{]} & $\Delta$MAE {[}\%{]} & $\Delta$RMSE {[}\%{]} & $\Delta$EH {[}\%{]} & $\Delta$NM {[}\%{]} \\
\midrule
1      & 3.20        & 5.42         & 8.20        & 68.56       & -38.70      & -41.09       & -27.29      & 10.13       \\
2      & 2.56        & 4.68         & 32.24       & 72.42       & -44.23      & -40.68       & -4.29       & 16.49       \\
3      & 2.50        & 4.70         & 35.91       & 72.86       & -45.53      & -40.73       & -1.25       & 16.79       \\
4      & 3.21        & 5.42         & 8.69        & 68.24       & -41.42      & -41.66       & -24.68      & 11.94       \\
5      & 2.75        & 4.95         & 9.87        & 76.17       & -36.05      & -40.86       & -36.55      & 8.76       \\
6      & 2.63        & 4.77         & 33.87       & 70.55       & -46.00      & -42.11       & 0.14        & 15.63       \\
7      & 2.59        & 4.60         & 24.96       & 70.93       & -38.48      & -36.11       & -11.10      & 15.34       \\
8      & 4.60        & 7.06         & 3.76        & 40.02       & -37.16      & -39.91       & -29.57      & -8.49       \\
9      & 1.24        & 1.86         & 35.16       & 86.16       & -19.48      & -33.09       & -19.52      & 11.90       \\
10     & 4.94        & 7.68         & 4.31        & 36.81       & -41.33      & -43.03       & -32.50      & -12.54       \\
\bottomrule
\end{tabular}
\label{tab:res_lr}
\end{table*}

The LR models reduce the MAE and RMSE errors in all test cases. For traffic lights, 1 -- 10, the MAE decreases between 1.24 and 3.20 sec. Only for traffic light 9 the improvement is lower with 19.48\%. Nevertheless, the MAE error of the naive model for this test case (1.24 sec) is already low compared to the other test cases. However, the error metrics still show high deviations from the test data set. The high variation of the T2G due to non-linear dependencies on the arrival of public transportation can not be captured by the LR models. Also, the results of the EH- and NM-ratio stresses the importance of this assessment: The EH-ratios decrease significantly for all traffic signals, except traffic signal 6. However, for that signal no substantial improvements can be observed (an increase in EH-ratio of 0.14\%). For the NM-ratios, the ratios improve between 8.76\% and 16.79\%. However, for traffic signal 8 and 10, the performance in NMs decreases by 8.49\% and 12.54\%, respectively. Note that only evaluating the LR models by the MAE would lead to an acceptable performance improvement compared to the naive model. Nevertheless, the EH- and NM ratios draw a different picture and stress the importance of these metrics. 

Figure~\ref{fig:result_comp2} (a) shows the test data with lower variation in the T2G test samples again. 
The LR model for traffic signal 4 can represent the general trend of the time series but can not predict high peaks of the T2G. In addition, the results underline the EH and NM ratios shown in Table~\ref{tab:res_lr}. For the time frames where the T2G remains constant, the LR model overestimates the T2G consistently. Figure~\ref{fig:result_comp2} (b) shows that the LR model for traffic signal 6 fails to reproduce the T2G pattern and also to predict high variations in the T2G (cycle 20 to 30). Nevertheless, the MAE and RMSE decrease for both cases (Table~\ref{tab:res_lr}). 

As third candidates, RF models are implemented for predicting the T2G. Again, we utilize the data set after the application of the RFE to guarantee a fair performance assessment. RF models require the tuning of a set of hyperparameters. In this work, we selected the following set of parameters and corresponding distributions for parameter sampling: the number of estimators, denoted as a uniformed distribution $\mathcal{U}(50, 200)$; max depth denoted as $\mathcal{U}(3,12)$; min samples split, denoted as $\mathcal{U}(2,6)$; and min weight fraction leaf as $\mathcal{U}(0, 0.5)$. The defined distributions specify the parameter search space for the TPE algorithm. Ten trail runs are executed, and the best model performance is obtained with the mean absolute error criterion. Note that the tuning procedure is applied to every model separately to maximize performance. A further increase of, e.g., the number of estimators or max depth of trees improves the performance only slightly, whereas the computational time to fit the model increases substantially. 

Table~\ref{tab:res_rf} shows the performance of the prediction models on the test data set for every traffic signal at the intersection. The MAE, RMSE, EH ratio, and NM ratio are again compared to the Naive model results from Table~\ref{tab:res_naive}. An assessment of the decision-tree-based method shows that all MAE values decrease by 41.75\% up to 61.16\% for traffic lights 1 -- 10. Note that this constitutes a reduction of MAE between 14.96\% and 34.18\% compared to the LR models. 

\begin{table*}[tb]
\centering
\caption{Model performance on the test data set with $\rm MAE$, $\rm RMSE$, $\mathrm{EH}$, and $\mathrm{NM}$ for the RF model. Also, the improvements over the naive baseline are presented with the given performance metrics.}
\begin{tabular}{rrrrr|rrrr}
\toprule
       & \multicolumn{4}{c}{RF model}                           & \multicolumn{4}{c}{Improvement over naive baseline}    \\
       \midrule
Signal & MAE {[}s{]} & RMSE {[}s{]} & EH {[}\%{]} & NM {[}\%{]} & $\Delta$MAE {[}\%{]} & $\Delta$RMSE {[}\%{]} & $\Delta$EH {[}\%{]} & $\Delta$NM {[}\%{]} \\
\midrule
1      & 2.23        & 5.57         & 62.77       & 80.81       & -57.38      & -39.46       & 27.28       & 22.38       \\
2      & 2.18        & 4.72         & 54.68       & 76.35       & -52.57      & -40.14       & 18.15       & 20.42       \\
3      & 2.08        & 4.71         & 58.56       & 77.17       & -54.64      & -40.62       & 21.40       & 21.10       \\
4      & 2.27        & 5.50         & 59.71       & 79.47       & -58.59      & -40.81       & 26.34       & 23.17       \\
5      & 1.81        & 5.13         & 69.71       & 84.65       & -57.91      & -38.71       & 23.29       & 17.24      \\
6      & 2.22        & 4.79         & 52.69       & 76.40       & -54.41      & -41.87       & 18.96       & 21.48       \\
7      & 2.10        & 4.42         & 51.21       & 75.69       & -50.02      & -38.55       & 15.15       & 20.10       \\
8      & 3.16        & 7.18         & 54.39       & 73.99       & -56.83      & -38.87       & 21.06       & 25.48       \\
9      & 0.90        & 2.02         & 68.16       & 86.25       & -41.75      & -27.23       & 13.48       & 11.99       \\
10     & 3.27        & 7.81         & 59.57       & 74.82       & -61.16      & -42.06       & 22.76       & 25.47       \\
\bottomrule
\end{tabular}
\label{tab:res_rf}
\end{table*}

\begin{figure}
    \centering
  \subfloat[]{%
       \includegraphics[width=0.8\linewidth]{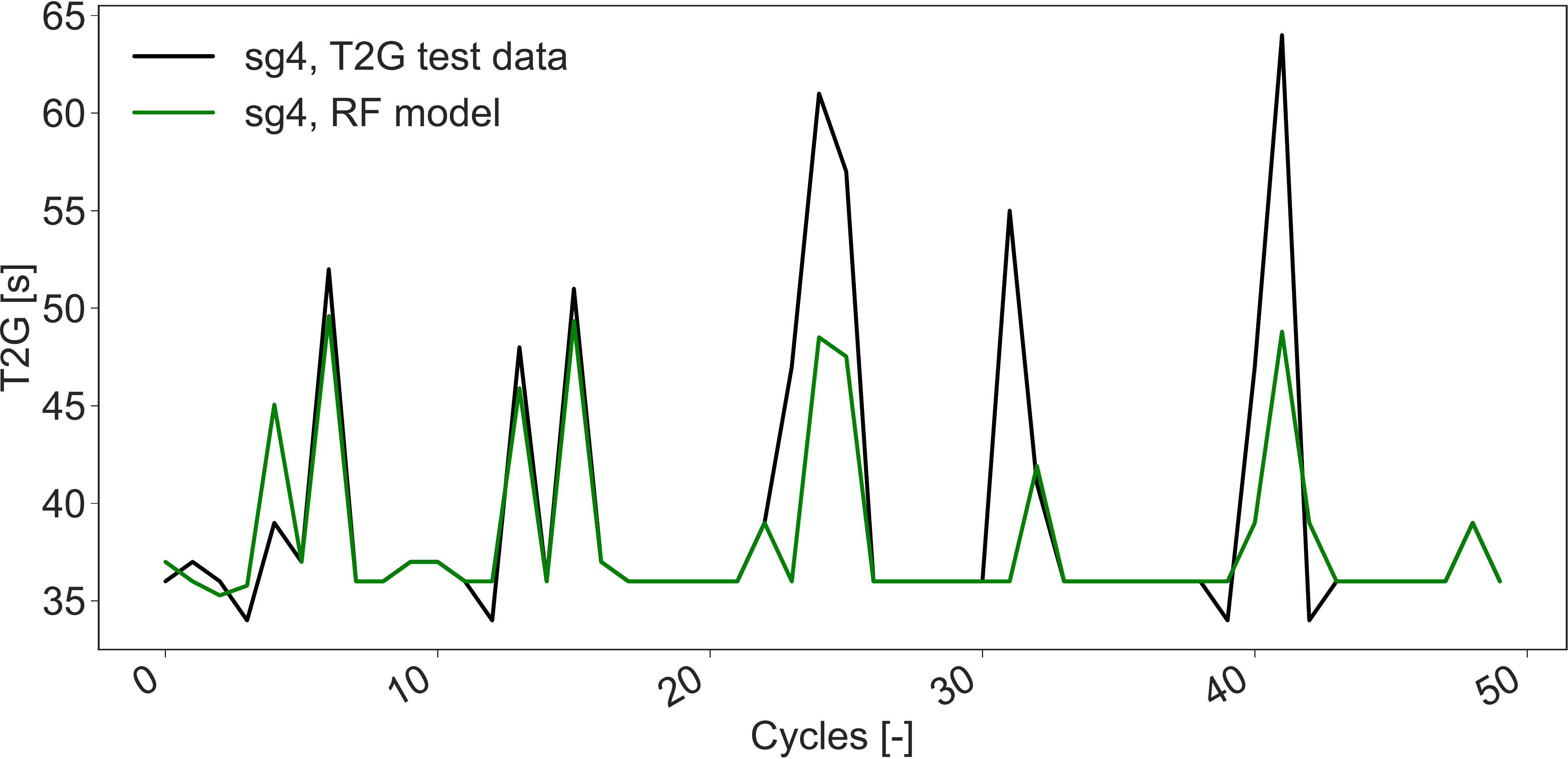}} \\
  \subfloat[]{%
        \includegraphics[width=0.8\linewidth]{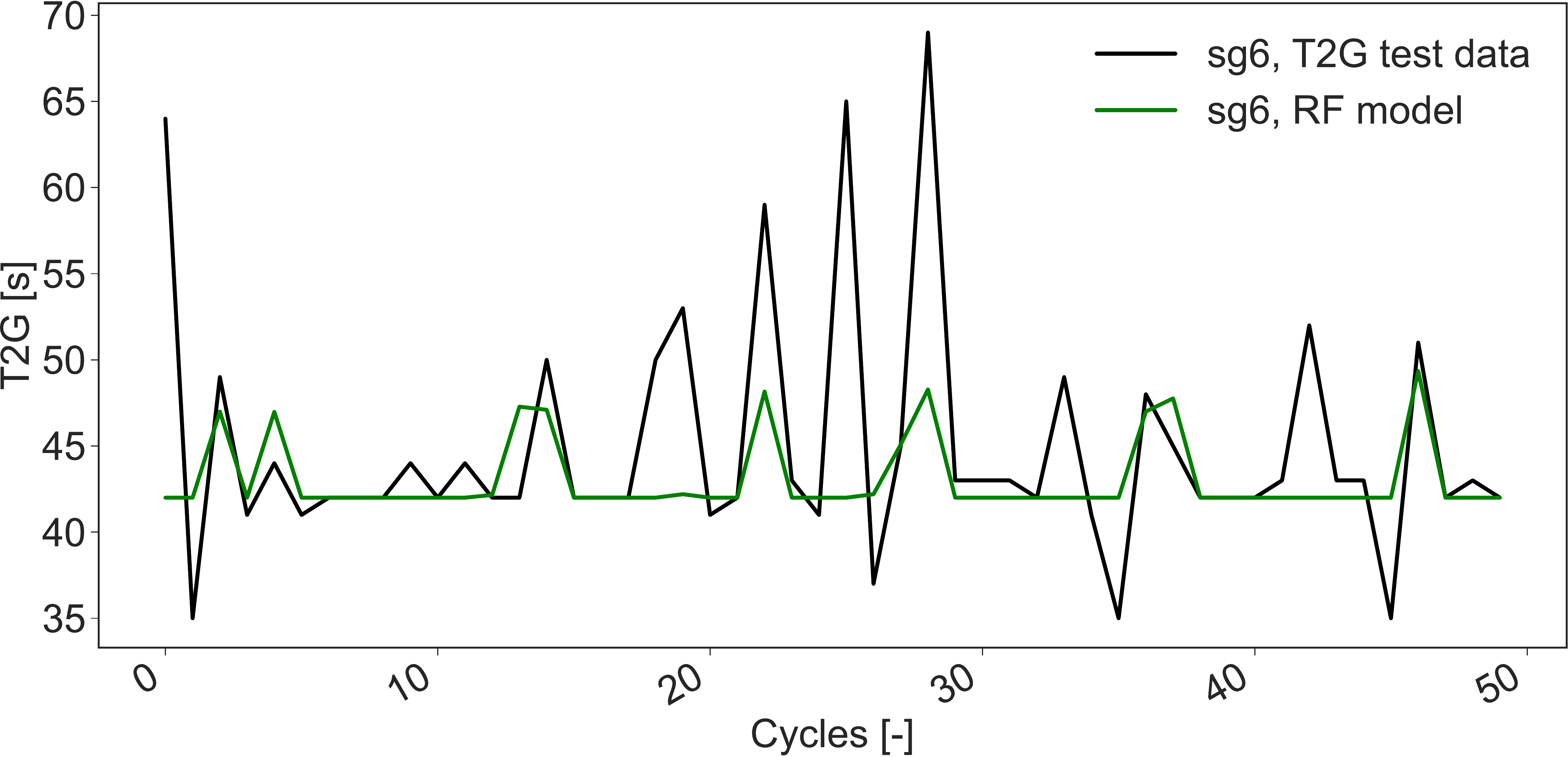}}
        
  \caption{Prediction results of the RF model. Results assessed on the T2G test data set for (a) traffic signal 4 and (b) traffic signal 6.}
  \label{fig:result_comp3} 
\end{figure}

The two samples of RF predictions for 50 cycles of the test data set are shown in Figure~\ref{fig:result_comp3}. The RF model in Figure~\ref{fig:result_comp3} shows a good fit when the signal control system is operating in its standard phase, and the time series trend can be replicated for traffic light 4 (overall MAE=2.27 sec; EH=59.71\%; NM=79.47\%). Note that the LR model overestimates the T2G within such time frames. Also, high variations up to a T2G of 50 sec can be predicted accurately. Nevertheless, higher peaks in the T2G samples (Figure~\ref{fig:result_comp3} (a) and (b)) can not be captured by any of the two models. Potential reasons for this failure are detections of public transportation that occur after the prediction time stamp of the next T2G. More details are discussed in Section~\ref{sec:diss}. 

Last, LSTM neural network models are implemented to predict the T2G. Again the data set after the application of the RFE is utilized. The application of the KerasTuner finds the architectures of the LSTM models by sampling the following parameters: the number of units for an LSTM layer, specified as $\mathcal{U}(8, 256)$; the number of hidden layers specified as $\mathcal{U}(1, 2)$; the dropout rate for regularization purposes specified as $\mathcal{U}(0, 0.5)$; the number of units for a dense layer connected after the LSTM layers, specified as $\mathcal{U}(1,50)$. The activation function is fixed to ReLU. For the training of the models, 100 epochs are computed with early stopping criteria by monitoring the validation loss with a patience $p=5$ is specified to prevent overfitting to the training data set. The batch size is set to 64, the validation split to 20\% of the training data set, and the MAE is utilized as a loss function. The considered time lag for the LSTM models is chosen with 7 time steps, determined with a sensitivity analysis. 

In Table~\ref{tab:res_lstm} the compilation of performance metrics for the LSTM models are shown. Overall, a reduction in MAE and RMSE compared to the Naive model can be achieved by deploying LSTM neural networks. Also, the EH and NM ratios are increased significantly and outperform the Naive models: The EH-ratio for the naive model ranges from 33.33\% to 54.68\%, whereas the LSTM models achieve performance metrics between 43.05\% and 68.80\%.  The LSTM models for traffic lights 1 -- 10 perform with an MAE error below 5.00 sec, and the EH-/NM-ratios are above 43\% and 61\%, respectively. 

The LSTM neural networks can not outperform the decision tree ensembles concerning all considered performance metrics compared with the RF results. For example, the MAE and RMSE errors when applying the LSTM models for traffic lights 1 -- 10 are between 12.60\% sec and 52.91\% higher. In addition, the EH and NM ratio decrease by -2\% up to -19\%. Only for the T2G predictions of traffic lights 7 and 9, the EH ratio is improved by 1.47\% and 0.64\%, respectively. This behavior can be explained by the low variability of the T2G target values for these traffic lights.

\begin{table*}[tb]
\centering
\caption{Model performance on the test data set with $\rm MAE$, $\rm RMSE$, $\mathrm{EH}$, and $\mathrm{NM}$ for the LSTM model. Also, the improvements over the naive baseline are presented with the given performance metrics.}
\begin{tabular}{rrrrr|rrrr}
\toprule
       & \multicolumn{4}{c}{LSTM model}                           & \multicolumn{4}{c}{Improvement over naive baseline}    \\
       \midrule
Signal & MAE {[}s{]} & RMSE {[}s{]} & EH {[}\%{]} & NM {[}\%{]} & $\Delta$MAE {[}\%{]} & $\Delta$RMSE {[}\%{]} & $\Delta$EH {[}\%{]} & $\Delta$NM {[}\%{]} \\
\midrule
1      & 3.15        & 6.54         & 43.69       & 72.94       & -39.66      & -28.91       & 8.20        & 14.51       \\
2      & 2.51        & 5.39         & 50.65       & 71.10       & -45.32      & -31.69       & 14.12       & 15.17       \\
3      & 2.59        & 5.47         & 53.19       & 71.45       & -43.57      & -31.02       & 16.03       & 15.38       \\
4      & 3.27        & 6.57         & 43.05       & 70.66       & -40.33      & -29.28       & 9.68        & 14.36       \\
5      & 2.48        & 6.30         & 64.81       & 81.65       & -42.33      & -24.73       & 18.39       & 14.24       \\
6      & 2.78        & 5.56         & 46.29       & 70.08       & -42.92      & -32.52       & 12.56       & 15.16       \\
7      & 2.44        & 4.96         & 52.68       & 70.41       & -42.04      & -31.11       & 16.62       & 14.82       \\
8      & 4.39        & 8.24         & 44.51       & 61.60       & -40.03      & -29.87       & 11.18       & 13.09       \\
9      & 1.01        & 2.19         & 68.80       & 83.86       & -34.42      & -21.22       & 14.12       & 9.60       \\
10     & 5.00        & 9.45         & 47.59       & 61.41       & -40.62      & -29.90       & 10.78       & 12.06       \\
\bottomrule
\end{tabular}
\label{tab:res_lstm}
\end{table*}

\begin{figure}
    \centering
  \subfloat[]{%
       \includegraphics[width=0.8\linewidth]{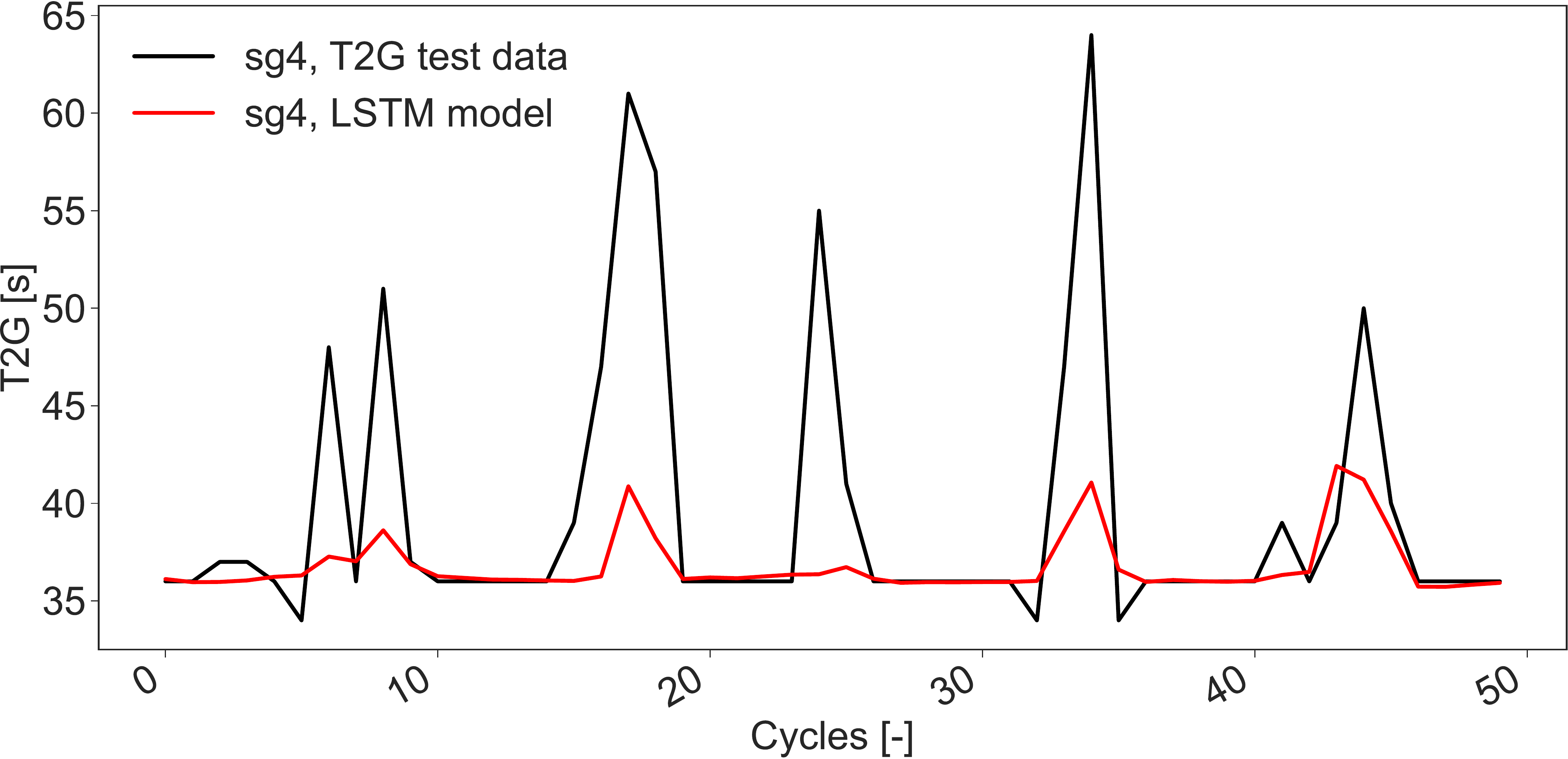}} \\
  \subfloat[]{%
        \includegraphics[width=0.8\linewidth]{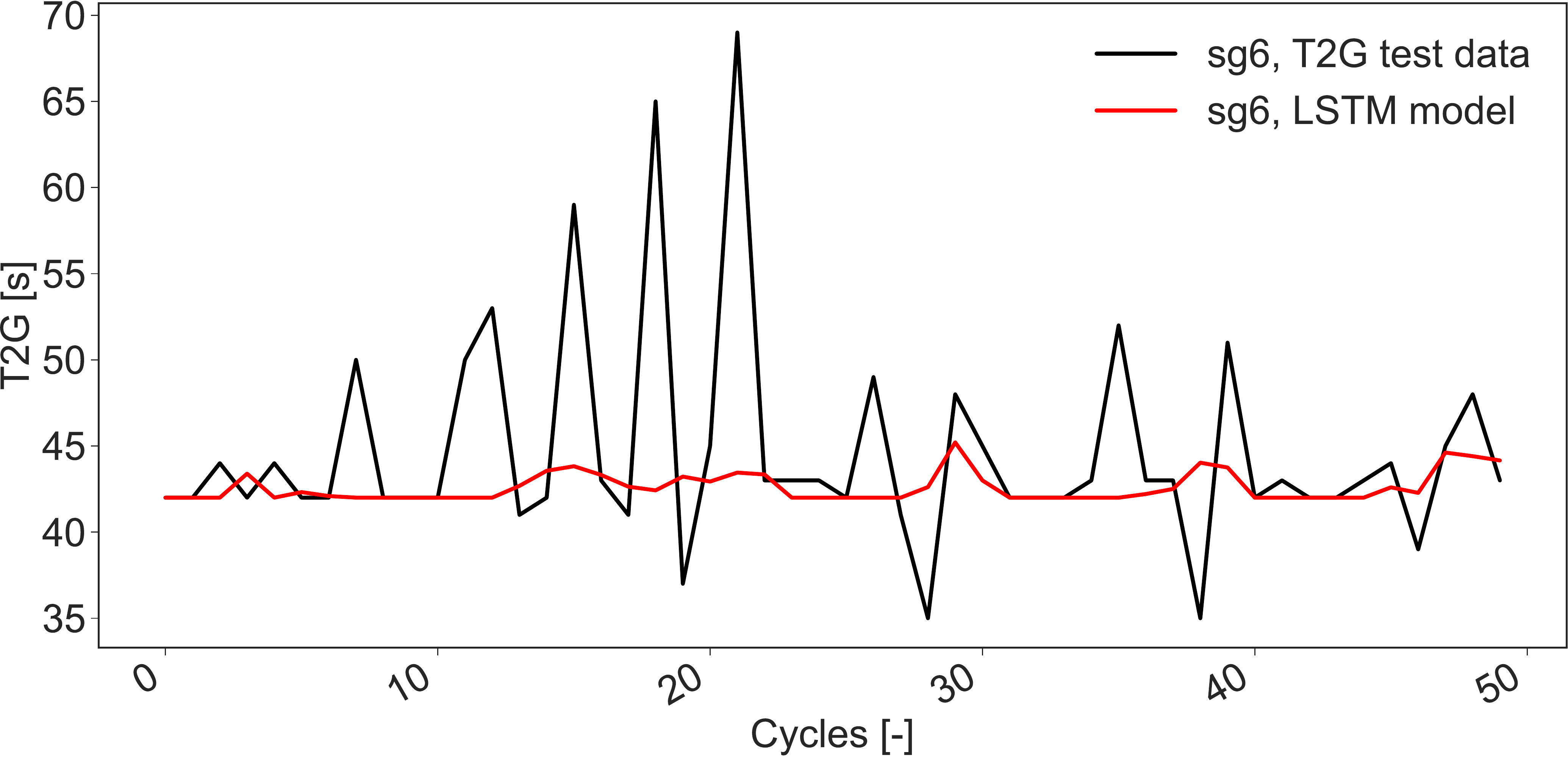}}
        
  \caption{Prediction results of the LSTM model. Results assessed on the T2G test data set for (a) traffic signal 4 and (b) traffic signal 6.}
  \label{fig:result_comp4} 
\end{figure}

Figure~\ref{fig:result_comp4} shows that the predicted T2G series for traffic lights 4 and 6 are converging towards the mean value of the corresponding T2G. Hence, an LSTM model that predicts the average T2G performs well on test data sets with low variability of the target variable. The predictions in Figure~\ref{fig:result_comp4} show in both cases that the LSTM models perform well in predicting the T2G when no high variability of the T2G is present. Nevertheless, none of the higher variations are not captured with these models, although the RF model predicted and approximated these samples with high accuracy. 
 
\begin{figure*}[tb]
   \centering
    \includegraphics[width=1\textwidth]{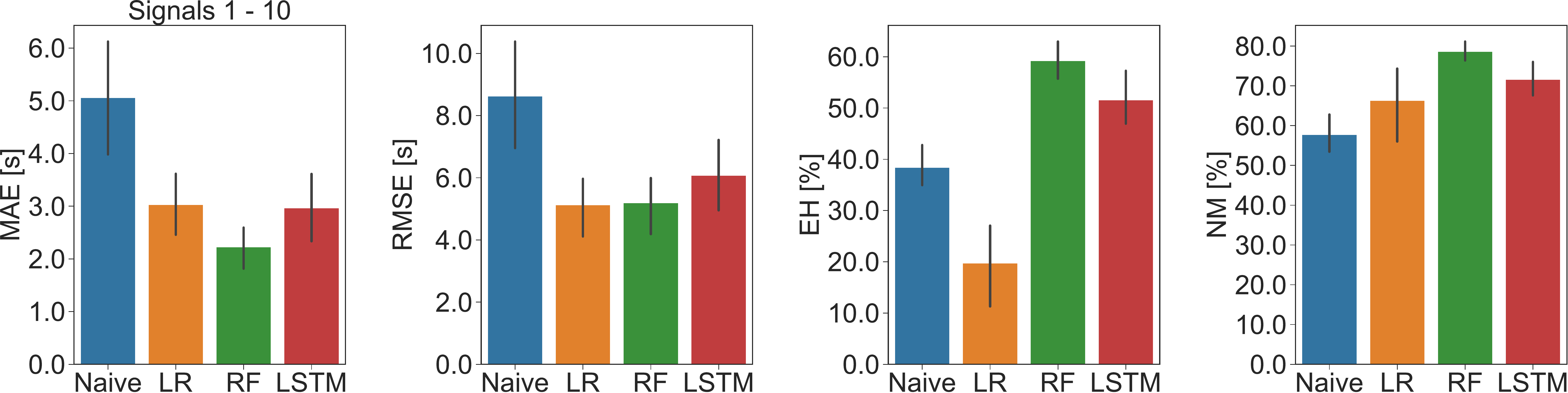}
  \caption{Performance of all signals 1--10 with respect to MAE, RMSE, EH and NM ratio.}\label{fig:comp_final_results}
\end{figure*}

Finally, we compare the results of all models in Figure~\ref{fig:comp_final_results}. The subplots show the MAE, RMSE, EH, and NM metrics for signals 1 -- 10. The RF models show the lowest error values with an MAE of 2.22 sec (standard deviation: 0.66 sec), an EH-ratio of 59.14\% (standard deviation: 6.27\%), and an NM-ratio of 78.56\% (standard deviation:  4.17\%). Note that the aggregated RMSE metrics of the LR and RF models show no significant difference with 5.11 and 5.18 sec, respectively. 

\subsection{Feature importance of Random Forest models}
\label{sec:feature_importance}
One advantage of the best candidate in this study is that RFs allow for a straightforward computation of the feature importance. One decision tree in a random forest splits input values based on the condition of impurity. When solving a regression problem, impurity is defined as the variance. When training the model, the weighted impurity should be minimized. Each feature's contribution allows for the calculation of the feature importance to solve the initial problem definition $\hat{Y}_{i}(c_{i,n}) = f(X)$; i.e., the approximation of a function that maps the input feature vector to the T2G target values. 

\begin{figure*}
    \centering
  \subfloat[]{%
       \includegraphics[width=0.5\linewidth]{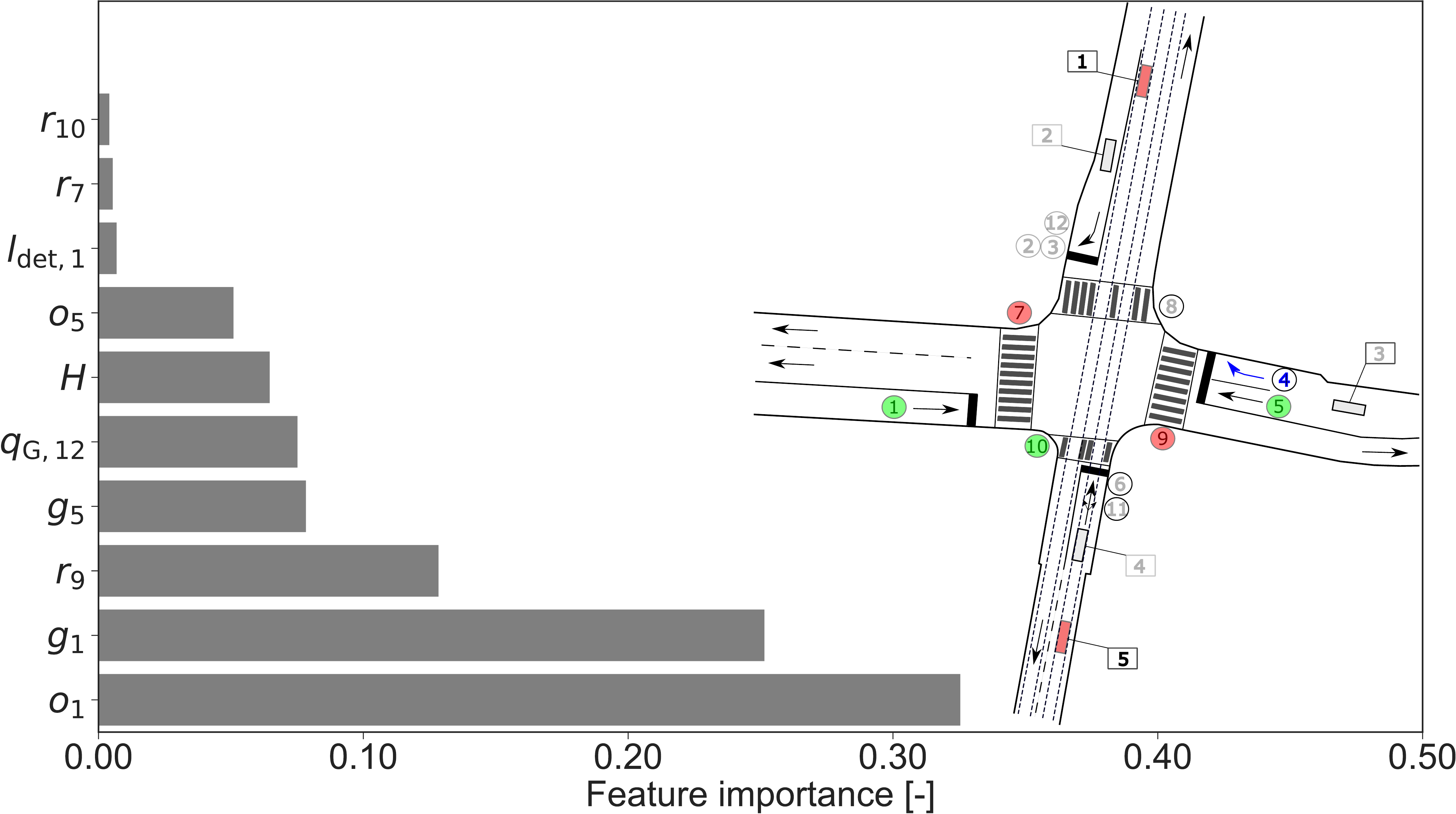}} 
  \subfloat[]{%
        \includegraphics[width=0.5\linewidth]{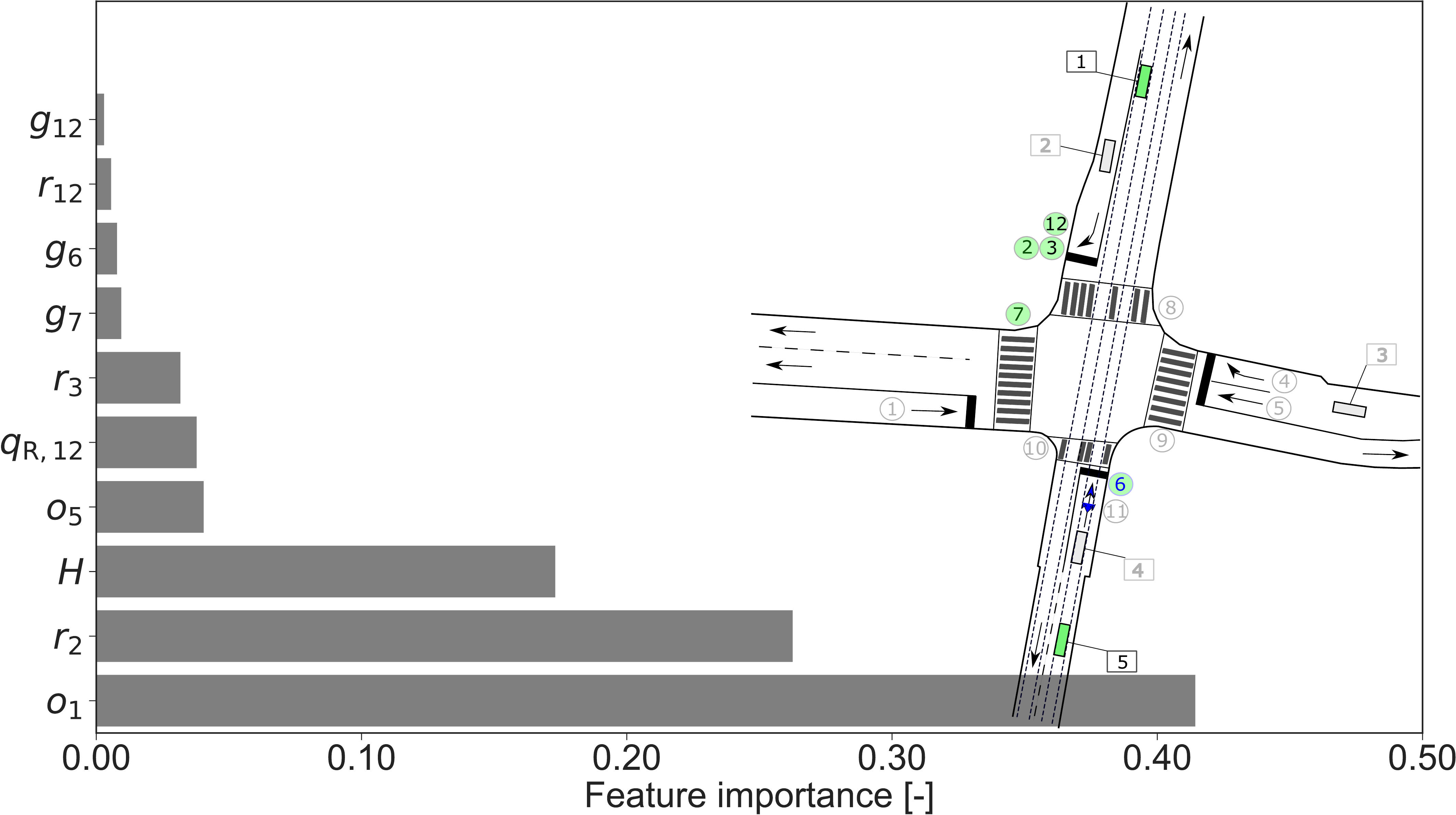}}
        
  \caption{Feature importance for RF model of (a) for traffic signal 4 and (b) traffic signal 6.}
  \label{fig:feature_importance} 
\end{figure*}

We compute the feature importance for the 10 most relevant features for the models of traffic signals 4 and 6, respectively. Note that other importance vectors can be computed analogously. Figure~\ref{fig:feature_importance} shows the 10 most important features for the prediction of the T2G of traffic signals 4 and 6. In the case of signal 4 $o_{1}$ is the most important, and $r_{10}$ is the least important feature of the presented subset (Figure~\ref{fig:feature_importance} (a)). Figure~\ref{fig:feature_importance} also highlights the feature importance on the intersection. The traffic stream in blue shows the signal for which the T2G prediction is computed. In red, the relevant devices (LDs or traffic signals) of conflicting traffic streams are shown, and green highlights the devices of the compatible traffic streams. 

For the both signals, the most important feature appears to be the occupancy of LD 1, i.e.,  $o_1$ detecting arriving trams from the north of the intersection area. This is expected as the T2G is highly dependent on the priority of public transportation. Additionally, for both models, the occupancy $o_5$ for arriving vehicles and trams from the south is listed in the 10 most relevant features. For the T2G prediction of traffic signal 4, the second most important feature is the green phase's duration of signal 1, $g_1$ (non-conflicting traffic stream); for the T2G of traffic signal 6, it is the red phase duration, $r_2$. Finally, note that in both cases, the feature representing the hour of the day $H$ is important and highlights that both models find T2G patterns that depend on the time of the day. 

Interesting all computed features introduced in Section~\ref{sec:methodology} appear at least once in one or the other feature importance subsets of the two presented models. On the other hand, the congestion and queue indicator $\mathrm{QI}_i(c_{i,n})$ and $\mathrm{CI}_i(c_{i,n})$ do not appear, and an analysis shows that the RFE procedure already eliminated these features. 

\section{DISCUSSION}
\label{sec:diss}

\subsection{Metrics for model evaluation}
\label{sec:model_evaluation}
A model for T2G predictions has to meet strong accuracy requirements. For example, motion planning algorithms of automated vehicles can utilize such forecasts. Hence, a low accuracy prediction of the following green can cause safety issues that are not acceptable in practice. Consequently, a judgment based on standard metrics such as the MAE or RMSE can lead to a good performance on average, but individual predictions might still not meet the initial requirements. Therefore, in this study, we introduced the EH and NM to evaluate models based on the forecast being identical to the target, or an error smaller or equal to two sec, respectively.

For example, the performance concerning the MAE for the LR models (Table~\ref{tab:res_lr}) shows errors that are close to the ones for the RF models (Table~\ref{tab:res_rf}). Nevertheless, the EH and NM ratio in Table~\ref{tab:res_rf} significantly improved compared to the naive baseline. On the contrary, the performance ratios of the LR models even decrease compared to the naive model. This highlights the importance of EH and NM for this problem. 

Although the hyperparameter tuning was carried out by assessing different loss functions, the MAE function showed the best performance concerning all presented performance metrics. For example, utilizing the mean squared error as a loss function did not improve the results. A more extensive data set might help improve generalization and performance. However, for the RF models, the loss function allows for the most accurate results. In addition, RF models are easier to fit and allow for interpretation of the model parameters (the feature importance analysis described in Section~\ref{sec:feature_importance}).

\subsection{Vehicle detection after T2G prediction}
\label{sec:late_detections}
As shown in Section~\ref{sec:feature_importance} the LD data representing trams' detection is of great importance for the RF models. However, results also show that models sometimes fail to predict a T2G peak (Figure~\ref{fig:result_comp3} (b)). One potential explanation for this behavior is detections of vehicles that occur after the prediction of the T2G. In other words, we predict the duration of the next red phase, and afterward, the corresponding phase starts. If a tram arrives at an intersection approach within this phase, the signal control system might react according to predefined conditions. As a result, the red phase can be shortened or extended (dependent on the traffic relation), and the T2G duration also changes. However, this information is only available in the next cycle. Therefore, the presented prediction models potentially miss high peaks of the T2G. 

A more extensive data set or additional feature engineering could help eliminate this limitation. Additionally, works such as~\cite{ref:Ibrahim} capture this system behavior as the predictions are updated consistently during the red phase. Nevertheless, this leads to fluctuations in the T2G that are problematic for control systems (e.g., motion planning of automated vehicles). Ibrahim et al.~\cite{ref:Ibrahim} also requires data aggregation per cycle length; meaning knowledge of the cycle length must be present a priori which is only possible for semi-actuated signal control systems.

\subsection{Prediction of T2G for dedicated public transportation signals}
\label{sec:PT_T2G_prediction}
As presented in Figure~\ref{fig:intersection}, traffic signals 11 and 12 are dedicated to public transportation vehicles. These traffic lights only operate in a green phase when a tram is detected and needs to pass the intersection. Hence, the average red and green times differ significantly from those of signals 1--10: For signals 11 and 12, the average red/green time are 203.12 sec/13 sec, and 214.50 sec/17.65 sec, respectively. Also, the minimum and maximum values range from 5 to 500 sec, 12 to 500 sec for the red times, and 6 to 353 sec, 4 to 320 sec for the green times. Note that the maximum allowed duration of 180 sec does not apply to these signals. Besides, the green times are significantly shorter, indicating that these traffic lights are only utilized when public transportation vehicles are detected. Consequently, a significantly higher variance of the quantities is given, which needs to be captured by the applied prediction model.
Also, the described limitation from Section~\ref{sec:late_detections} that vehicle detections after the T2G prediction, i.e., occurring within the red phase we predicted the duration for, has a substantial influence on model performance. As signals 11 and 12 only operate in a green phase when a vehicle is detected, it is evident that many detections occur during a red phase.

\begin{figure}[tb]
   \centering
    \includegraphics[width=0.5\textwidth]{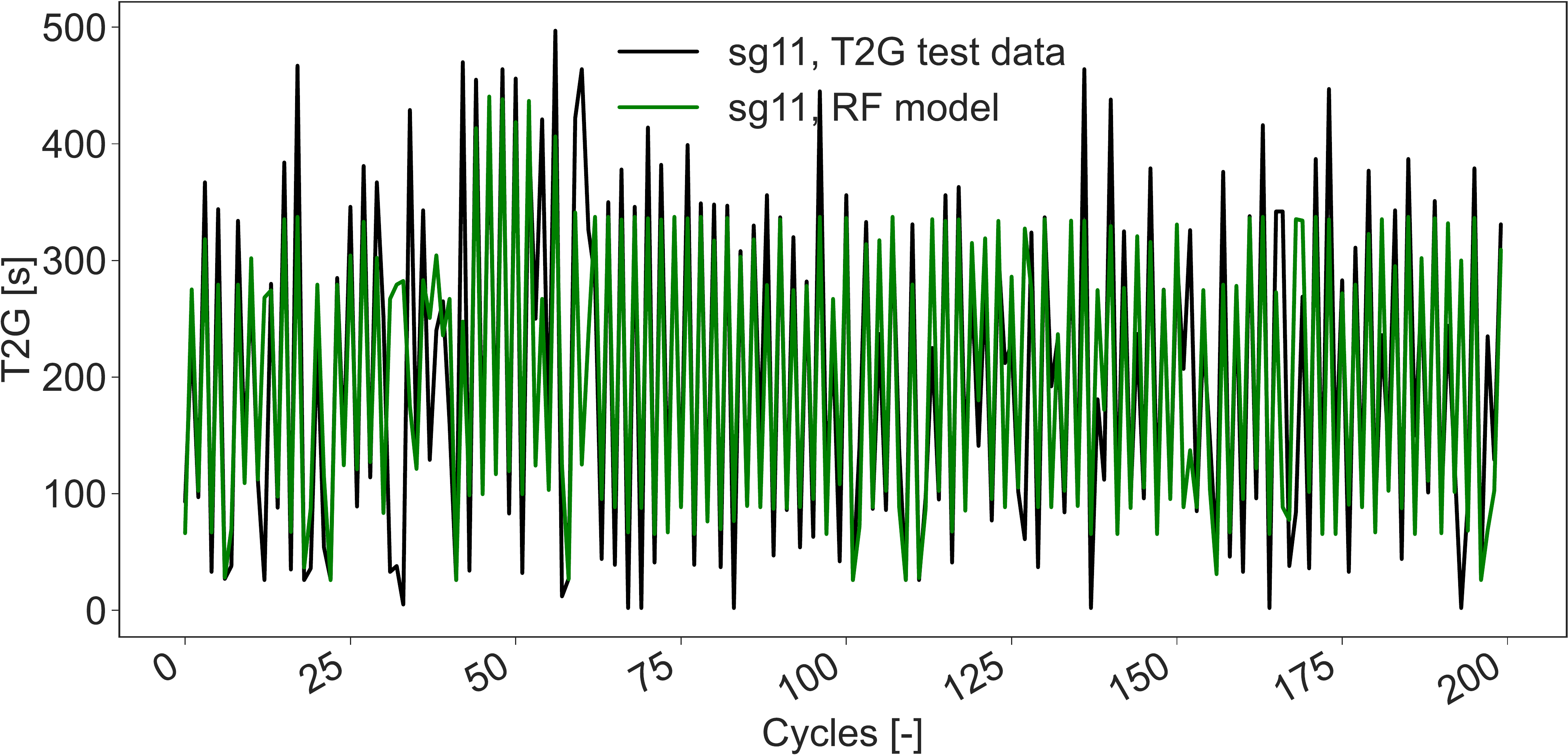}
  \caption{Prediction results of the RF model for traffic signal 11 only utilized by public transportation.}\label{fig:sig12_rf_pred}
\end{figure}

In Figure~\ref{fig:sig12_rf_pred} we shortly present the prediction results of the RF model for 200 cycles of traffic signal 11. 

The test data in black shows high fluctuations between 5 and 500 seconds that correspond to the min and max values found in the descriptive analysis. The RF model captures the time series pattern with most prediction samples. Nevertheless, the prediction errors are high compared to the values shown for signals 1--10. Table~\ref{tab:res_rf_11_12} compiles the MAE, RMSE, EH, and NM ratio for the RF models. 

\begin{table}[!t]
\centering
\caption{Model performance on the test data set with $\rm MAE$, $\rm RMSE$, $\mathrm{EH}$, and $\mathrm{NM}$ for the RF model on signal 11 and 12.}
\begin{tabular}{rrrrr}
\toprule
       & \multicolumn{4}{c}{RF model}           \\
       \midrule
Signal & MAE {[}s{]} & RMSE {[}s{]} & EH {[}\%{]} & NM {[}\%{]} \\
\midrule
11      & 61.56       & 92.40        & 3.21        & 8.12             \\
12     & 54.08       & 78.61        & 3.87        & 8.33       \\       
\bottomrule
\end{tabular}
\label{tab:res_rf_11_12}
\end{table}

As expected, the performance metrics show a significantly higher magnitude with an MAE of 85.88 and 64.02 sec, respectively. This is because the RF models can not capture the high T2G peaks that frequently occur for signals 11 and 12. Additionally, the EH and NM ratios underline the modest performance with values below 2\% for all metrics in Table~\ref{tab:res_rf_11_12}. 

A model design for designated traffic lights that satisfies accuracy requirements, additional information such as GPS signals that provide the location of trams, or additional LDs that allow inferring location or speed is needed.

\subsection{Utilization of T2G predictions for speed-advisory systems}
\label{sec:t2g_predictions_utilization}
T2G predictions can serve as an input to motion planning algorithms leading to a smoother speed profile and more homogeneous traffic flow as vehicles do not need to stop at an intersection. Our proposed methodology considers the complex relationship between traffic signals and LD detections to determine the T2G. Nevertheless, when motion planning algorithms consider the predictions of the T2G for determining, e.g., the speed profile to cross an intersection, the timestamp of an LD detection will also change. Consequently, the proposed models in this work that learned this temporal relationship via a historical data set (offline learning) can not dynamically adapt to the new system behavior. As a solution, the authors suggest researching the directions of (a) online learning, which allows for learning new data patterns as they are available, or (b) meta-learning allowing ML models to learn from T2G prediction outputs and adapt to the new system behavior.  

\section{CONCLUSION}
\label{sec:diss_concl}
This paper proposes a framework for Time-to-Green (T2G) predictions at an urban intersection to enhance the quality of Signal Phase and Timing (SPaT) messages. The problem was constructed as a time series forecast to predict the next signal phase of a fully-actuated signal control system. The framework implementation is generic and can be applied to any intersection that provides Loop Detectors (LD) and signals data. An extensive feature engineering methodology is proposed to enhance the model quality by utilizing concepts from traffic flow theory. To assess the performance of supervised learning algorithms, a Linear Regression (LR), a Random Forest (RF), and a Long-Short-Term-Memory (LSTM) neural network are implemented and assessed with a set of performance metrics. 

In the presented numerical experiment, the methodology was tested on an intersection in Zurich operated by a fully-actuated signal control and public transport priority. A consecutive data set of two months (traffic signal and LD data) is processed, and prediction models are assessed on the accuracy when predicting the T2G. Results show that RF models are promising tools for predicting the next red phase and outperform naive baseline, LR, and LSTM models with Mean Absolute Errors (MAE) between 0.90 and 3.27 sec. Nevertheless, the RF models show limitations in predicting the T2G of traffic lights designated for public transportation due to the high variance of the target values and vehicle detections after prediction. Future work will extend the present research with the possibility of updating T2G predictions throughout the next signal phase that can serve as an input to various control systems. We will also look at the parameter tuning of the models  concerning computational time. This paves the way for real-time applications. Another promising direction is to develop an algorithm that can be quickly adapted to new environments (e.g., other intersections or scenarios with different transit operations) within a few shots via meta-learning (e.g., \cite{gammelli2022graph}). 

\addtolength{\textheight}{0cm}   





\section*{ACKNOWLEDGMENT}
The authors would like to thank Felix Denzler from the city of Zurich for collaborating in a research project and to provide the data. K. Yang acknowledges the support by the Ministry of Education, Singapore, under its Academic Research Fund Tier 1. 
M. Menendez acknowledges the support by the NYUAD Center for Interacting Urban Networks (CITIES) funded by Tamkeen under the NYUAD Research Institute Award CG001. L. Ambühl acknowledges the support by ETH Research Grant ETH-27 16-1 under the name of SPEED.


\addtolength{\textheight}{0.2cm}
\bibliographystyle{IEEEtran}
\bibliography{IEEE_T_ITS}

\begin{thebibliography}{10}
\providecommand{\url}[1]{#1}
\csname url@samestyle\endcsname
\providecommand{\newblock}{\relax}
\providecommand{\bibinfo}[2]{#2}
\providecommand{\BIBentrySTDinterwordspacing}{\spaceskip=0pt\relax}
\providecommand{\BIBentryALTinterwordstretchfactor}{4}
\providecommand{\BIBentryALTinterwordspacing}{\spaceskip=\fontdimen2\font plus
\BIBentryALTinterwordstretchfactor\fontdimen3\font minus
  \fontdimen4\font\relax}
\providecommand{\BIBforeignlanguage}[2]{{%
\expandafter\ifx\csname l@#1\endcsname\relax
\typeout{** WARNING: IEEEtran.bst: No hyphenation pattern has been}%
\typeout{** loaded for the language `#1'. Using the pattern for}%
\typeout{** the default language instead.}%
\else
\language=\csname l@#1\endcsname
\fi
#2}}
\providecommand{\BIBdecl}{\relax}
\BIBdecl

\bibitem{ref:claussmann_motion_planning}
L.~Claussmann, M.~Revilloud, D.~Gruyer, and S.~Glaser, ``A review of motion
  planning for highway autonomous driving,'' \emph{IEEE Transactions on
  Intelligent Transportation Systems}, vol.~21, no.~5, pp. 1826--1848, 2020.

\bibitem{ref:ZHENG2013_actuated_SC}
X.~Zheng and W.~Recker, ``An adaptive control algorithm for traffic-actuated
  signals,'' \emph{Transportation Research Part C: Emerging Technologies},
  vol.~30, pp. 93--115, 2013.

\bibitem{ref:laemmer_2008}
S.~Lämmer and D.~Helbing, ``Self-control of traffic lights and vehicle flows
  in urban road networks,'' \emph{Journal of Statistical Mechanics: Theory and
  Experiment}, vol. 2008, no.~04, p. P04019, apr 2008.

\bibitem{ref:genser_t2g_itsc}
A.~Genser, L.~Ambühl, K.~Yang, M.~Menendez, and A.~Kouvelas, ``Time-to-green
  predictions: A framework to enhance spat messages using machine learning,''
  in \emph{2020 IEEE 23rd International Conference on Intelligent
  Transportation Systems (ITSC)}, 2020, pp. 1--6.

\bibitem{ref:c2c}
\BIBentryALTinterwordspacing
{CAR 2 CAR Communication Consortium}, ``Automotive requirements for spat and
  map - car 2 car.'' [Online]. Available:
  \url{https://www.car-2-car.org/fileadmin/documents/Basic_System_Profile/Release_1.5.1/C2CCC_RS_2077_SPATMAP_AutomotiveRequirements.pdf}
\BIBentrySTDinterwordspacing

\bibitem{ref:genser_comfort_cacaie}
A.~Genser, R.~Spielhofer, P.~Nitsche, and A.~Kouvelas, ``Ride comfort
  assessment for automated vehicles utilizing a road surface model and monte
  carlo simulations,'' \emph{Computer-Aided Civil and Infrastructure
  Engineering}, 2021.

\bibitem{ref:fayazi1}
S.~A. {Fayazi}, A.~{Vahidi}, G.~{Mahler}, and A.~{Winckler}, ``Traffic signal
  phase and timing estimation from low-frequency transit bus data,'' \emph{IEEE
  Transactions on Intelligent Transportation Systems}, vol.~16, no.~1, pp.
  19--28, 2015.

\bibitem{ref:fayazi2}
S.~A. {Fayazi} and A.~{Vahidi}, ``Crowdsourcing phase and timing of pre-timed
  traffic signals in the presence of queues: Algorithms and back-end system
  architecture,'' \emph{IEEE Transactions on Intelligent Transportation
  Systems}, vol.~17, no.~3, pp. 870--881, 2016.

\bibitem{ref:wang}
C.~{Wang} and S.~{Jiang}, ``Traffic signal phases' estimation by floating car
  data,'' in \emph{2012 12th International Conference on ITS
  Telecommunications}, 2012, pp. 568--573.

\bibitem{ref:Yu}
{Juan Yu} and {Pei-zhong Lu}, ``Learning traffic signal phase and timing
  information from low-sampling rate taxi gps trajectories,''
  \emph{{Knowl.-Based Syst.}}, vol. 110, pp. 275--292, 2016.

\bibitem{ref:ban}
X.~J. Ban, R.~Herring, P.~Hao, and A.~M. Bayen, ``Delay pattern estimation for
  signalized intersections using sampled travel times,'' \emph{Transportation
  Research Record}, vol. 2130, no.~1, pp. 109--119, 2009.

\bibitem{ref:protschky}
V.~{Protschky}, C.~{Ruhhammer}, and S.~{Feit}, ``Learning traffic light
  parameters with floating car data,'' in \emph{2015 IEEE 18th International
  Conference on Intelligent Transportation Systems}, 2015, pp. 2438--2443.

\bibitem{ref:protschky3}
V.~Protschky, S.~Feit, and C.~Linnhoff-Popien, ``Extensive traffic light
  prediction under real-world conditions,'' in \emph{2014 IEEE 80th Vehicular
  Technology Conference (VTC2014-Fall)}, 2014, pp. 1--5.

\bibitem{ref:Ibrahim}
S.~{Ibrahim}, D.~{Kalathil}, R.~O. {Sanchez}, and P.~{Varaiya}, ``Estimating
  phase duration for spat messages,'' \emph{IEEE Transactions on Intelligent
  Transportation Systems}, vol.~20, no.~7, pp. 2668--2676, 2019.

\bibitem{ref:zhu}
Y.~Zhu, X.~Liu, M.~Li, and Q.~Zhang, ``Pova: Traffic light sensing with probe
  vehicles,'' \emph{IEEE Transactions on Parallel and Distributed Systems},
  vol.~24, no.~7, pp. 1390--1400, 2013.

\bibitem{ref:protschky2}
V.~{Protschky}, K.~{Wiesner}, and S.~{Feit}, ``Adaptive traffic light
  prediction via kalman filtering,'' in \emph{2014 IEEE Intelligent Vehicles
  Symposium Proceedings}, 2014, pp. 151--157.

\bibitem{ref:islam_cnn_lstm_spat}
Z.~Islam, M.~Abdel-Aty, and N.~Mahmoud, ``Using cnn-lstm to predict signal
  phasing and timing aided by high-resolution detector data,''
  \emph{Transportation Research Part C: Emerging Technologies}, vol. 141, p.
  103742, 2022.

\bibitem{ref:Wang_LSTM}
Y.~Wang, D.~Zhang, Y.~Liu, B.~Dai, and L.~H. Lee, ``Enhancing transportation
  systems via deep learning: A survey,'' \emph{Transportation Research Part C:
  Emerging Technologies}, vol.~99, pp. 144 -- 163, 2019.

\bibitem{ref:zhang}
J.~Zhang, F.-Y. Wang, K.~Wang, W.-H. Lin, X.~Xu, and C.~Chen, ``Data-driven
  intelligent transportation systems: A survey,'' \emph{IEEE Transactions on
  Intelligent Transportation Systems}, vol.~12, no.~4, pp. 1624--1639, 2011.

\bibitem{ref:chang}
C.~H., L.~Y., Y.~B, and B.~S., ``Dynamic near-term traffic flow prediction:
  system-oriented approach based on past experiences,'' \emph{IET Intelligent
  Transport Systems}, vol.~6, no.~3, pp. 292--305, 2012.

\bibitem{ref:sun}
H.~Sun, H.~X. Liu, H.~Xiao, R.~R. He, and B.~Ran, ``Use of local linear
  regression model for short-term traffic forecasting,'' \emph{Transportation
  Research Record}, vol. 1836, no.~1, pp. 143--150, 2003.

\bibitem{ref:clark}
S.~Clark, ``Traffic prediction using multivariate nonparametric regression,''
  \emph{Journal of Transportation Engineering}, vol. 129, no.~2, pp. 161--168,
  2003.

\bibitem{ref:genser}
A.~Genser, N.~Hautle, M.~Makridis, and A.~Kouvelas, ``An experimental urban
  case study with various data sources and a model for traffic estimation,''
  \emph{Sensors}, vol.~22, no.~1, p. 144, Dec 2021.

\bibitem{ref:hong}
W.-C. Hong, ``Traffic flow forecasting by seasonal svr with chaotic simulated
  annealing algorithm,'' \emph{Neurocomputing}, vol.~74, no. 12–13, 2011.

\bibitem{ref:castro_neto}
M.~Castro-Neto, Y.-S. Jeong, M.-K. Jeong, and L.~D. Han, ``Online-svr for
  short-term traffic flow prediction under typical and atypical traffic
  conditions,'' \emph{Expert Systems with Applications}, vol.~36, no. 3, Part
  2, pp. 6164--6173, 2009.

\bibitem{ref:jeong}
Y.-S. Jeong, Y.-J. Byon, M.~M. Castro-Neto, and S.~M. Easa, ``Supervised
  weighting-online learning algorithm for short-term traffic flow prediction,''
  \emph{IEEE Transactions on Intelligent Transportation Systems}, vol.~14,
  no.~4, pp. 1700--1707, 2013.

\bibitem{ref:zhang_RSF}
Y.~Zhang and A.~Haghani, ``A gradient boosting method to improve travel time
  prediction,'' \emph{Transportation Research Part C: Emerging Technologies},
  vol.~58, pp. 308--324, 2015, big Data in Transportation and Traffic
  Engineering.

\bibitem{ref:vlahogianni}
E.~I. Vlahogianni, M.~G. Karlaftis, and J.~C. Golias, ``Optimized and
  meta-optimized neural networks for short-term traffic flow prediction: A
  genetic approach,'' \emph{Transportation Research Part C: Emerging
  Technologies}, vol.~13, no.~3, pp. 211--234, 2005.

\bibitem{ref:chan}
K.~Y. Chan, T.~S. Dillon, J.~Singh, and E.~Chang, ``Neural-network-based models
  for short-term traffic flow forecasting using a hybrid exponential smoothing
  and levenberg–marquardt algorithm,'' \emph{IEEE Transactions on Intelligent
  Transportation Systems}, vol.~13, no.~2, pp. 644--654, 2012.

\bibitem{ref:kumar}
K.~Kumar, M.~Parida, and V.~Katiyar, ``Short term traffic flow prediction for a
  non urban highway using artificial neural network,'' \emph{Procedia - Social
  and Behavioral Sciences}, vol. 104, pp. 755--764, 2013, 2nd Conference of
  Transportation Research Group of India (2nd CTRG).

\bibitem{ref:tang}
J.~Tang, F.~Liu, Y.~Zou, W.~Zhang, and Y.~Wang, ``An improved fuzzy neural
  network for traffic speed prediction considering periodic characteristic,''
  \emph{IEEE Transactions on Intelligent Transportation Systems}, vol.~18,
  no.~9, pp. 2340--2350, 2017.

\bibitem{ref:lv}
Y.~Lv, Y.~Duan, W.~Kang, Z.~Li, and F.-Y. Wang, ``Traffic flow prediction with
  big data: A deep learning approach,'' \emph{IEEE Transactions on Intelligent
  Transportation Systems}, vol.~16, no.~2, pp. 865--873, 2015.

\bibitem{ref:huang}
W.~Huang, G.~Song, H.~Hong, and K.~Xie, ``Deep architecture for traffic flow
  prediction: Deep belief networks with multitask learning,'' \emph{IEEE
  Transactions on Intelligent Transportation Systems}, vol.~15, no.~5, pp.
  2191--2201, 2014.

\bibitem{ref:yu_rose}
R.~Yu, Y.~Li, C.~Shahabi, U.~Demiryurek, and Y.~Liu, \emph{Deep Learning: A
  Generic Approach for Extreme Condition Traffic Forecasting}, 2017, pp.
  777--785.

\bibitem{ref:qian}
F.~Qian, G.~Hu, and J.~Xie, ``A recurrent neural network approach to traffic
  matrix tracking using partial measurements,'' in \emph{2008 3rd IEEE
  Conference on Industrial Electronics and Applications}, 2008, pp. 1640--1643.

\bibitem{ref:koesdwiady}
A.~Koesdwiady, R.~Soua, and F.~Karray, ``Improving traffic flow prediction with
  weather information in connected cars: A deep learning approach,'' \emph{IEEE
  Transactions on Vehicular Technology}, vol.~65, no.~12, pp. 9508--9517, 2016.

\bibitem{ref:panwai}
S.~Panwai and H.~Dia, ``Comparative evaluation of microscopic car-following
  behavior,'' \emph{IEEE Transactions on Intelligent Transportation Systems},
  vol.~6, no.~3, pp. 314--325, 2005.

\bibitem{ref:ciuffo}
B.~Ciuffo, V.~Punzo, and M.~Montanino, ``Thirty years of gipps’ car-following
  model: Applications, developments, and new features,'' \emph{Transportation
  Research Record}, vol. 2315, no.~1, pp. 89--99, 2012.

\bibitem{ref:kesting}
A.~Kesting, M.~Treiber, M.~Schönhof, and D.~Helbing, ``Adaptive cruise control
  design for active congestion avoidance,'' \emph{Transportation Research Part
  C: Emerging Technologies}, vol.~16, no.~6, pp. 668--683, 2008.

\bibitem{ref:tran}
D.~Tran, W.~Sheng, L.~Liu, and M.~Liu, ``A hidden markov model based driver
  intention prediction system,'' in \emph{2015 IEEE International Conference on
  Cyber Technology in Automation, Control, and Intelligent Systems (CYBER)},
  2015, pp. 115--120.

\bibitem{ref:kim}
P.~J. Kim~IH, Bong~JH and P.~S., ``Prediction of driver's intention of lane
  change by augmenting sensor information using machine learning techniques,''
  \emph{Sensors}, vol. 17(6):1350, 2017.

\bibitem{ref:kumar_2}
P.~Kumar, M.~Perrollaz, S.~Lefèvre, and C.~Laugier, ``Learning-based approach
  for online lane change intention prediction,'' in \emph{2013 IEEE Intelligent
  Vehicles Symposium (IV)}, 2013, pp. 797--802.

\bibitem{ref:gurghian}
A.~Gurghian, T.~Koduri, S.~V. Bailur, K.~J. Carey, and V.~N. Murali,
  ``Deeplanes: End-to-end lane position estimation using deep neural
  networks,'' in \emph{2016 IEEE Conference on Computer Vision and Pattern
  Recognition Workshops (CVPRW)}, 2016, pp. 38--45.

\bibitem{ref:oluwatobi}
O.~Olabiyi, E.~Martinson, V.~Chintalapudi, and R.~Guo, ``Driver action
  prediction using deep (bidirectional) recurrent neural network,'' 2017.

\bibitem{ref:wang_car_following}
X.~Wang, R.~Jiang, L.~Li, Y.~Lin, X.~Zheng, and F.-Y. Wang, ``Capturing
  car-following behaviors by deep learning,'' \emph{IEEE Transactions on
  Intelligent Transportation Systems}, vol.~19, no.~3, pp. 910--920, 2018.

\bibitem{ref:riedel_2019}
T.~Riedel and M.~Menendez, ``Switzerland,'' in \emph{Global practices on road
  traffic signal control: Fixed-time control at isolated intersections},
  K.~Tang, M., Boltze, H.~Nakamura, and Z.~Tian, Eds.\hskip 1em plus 0.5em
  minus 0.4em\relax Elsevier, 2019, ch.~7, pp. 99--116.

\bibitem{ref:genser2020enhancement}
A.~Genser, L.~Amb{\"u}hl, K.~Yang, M.~Menendez, and A.~Kouvelas, ``Enhancement
  of spat-messages with machine learning based time-to-green predictions,'' in
  \emph{9th Symposium of the European Association for Research in
  Transportation (hEART 2020)}.\hskip 1em plus 0.5em minus 0.4em\relax European
  Association for Research in Transportation, 2020.

\bibitem{ref:tan_naive_traffic_flow}
M.-C. Tan, S.~C. Wong, J.-M. Xu, Z.-R. Guan, and P.~Zhang, ``An aggregation
  approach to short-term traffic flow prediction,'' \emph{IEEE Transactions on
  Intelligent Transportation Systems}, vol.~10, no.~1, pp. 60--69, 2009.

\bibitem{ref:smith_models_flow_forecast}
B.~L. Smith, B.~M. Williams, and R.~{Keith Oswald}, ``Comparison of parametric
  and nonparametric models for traffic flow forecasting,'' \emph{Transportation
  Research Part C: Emerging Technologies}, vol.~10, no.~4, pp. 303--321, 2002.

\bibitem{ref:mclaughlin1983forecasting}
R.~L. McLaughlin, ``Forecasting models: Sophisticated or naive?'' \emph{Journal
  of Forecasting (pre-1986)}, vol.~2, no.~3, p. 274, 1983.

\bibitem{ref:scikit-learn}
F.~Pedregosa, G.~Varoquaux, A.~Gramfort, V.~Michel, B.~Thirion, O.~Grisel,
  M.~Blondel, P.~Prettenhofer, R.~Weiss, V.~Dubourg, J.~Vanderplas, A.~Passos,
  D.~Cournapeau, M.~Brucher, M.~Perrot, and E.~Duchesnay, ``Scikit-learn:
  Machine learning in {P}ython,'' \emph{Journal of Machine Learning Research},
  vol.~12, pp. 2825--2830, 2011.

\bibitem{ref:LR1}
D.~Branston and H.~van Zuylen, ``The estimation of saturation flow, effective
  green time and passenger car equivalents at traffic signals by multiple
  linear regression,'' \emph{Transportation Research}, vol.~12, no.~1, pp. 47
  -- 53, 1978.

\bibitem{ref:breiman2001random}
L.~Breiman, ``Random forests,'' \emph{Machine learning}, vol.~45, no.~1, pp.
  5--32, 2001.

\bibitem{ref:hochreiter_LSTM}
S.~Hochreiter and J.~Schmidhuber, ``Long short-term memory,'' \emph{Neural
  computation}, vol.~9, no.~8, pp. 1735--1780, 1997.

\bibitem{ref:Ma_LSTM}
X.~Ma, Z.~Tao, Y.~Wang, H.~Yu, and Y.~Wang, ``Long short-term memory neural
  network for traffic speed prediction using remote microwave sensor data,''
  \emph{Transportation Research Part C: Emerging Technologies}, vol.~54, pp.
  187 -- 197, 2015.

\bibitem{ref:tensorflow}
M.~Abadi, P.~Barham, J.~Chen, Z.~Chen, A.~Davis, J.~Dean, M.~Devin,
  S.~Ghemawat, G.~Irving, M.~Isard, M.~Kudlur, J.~Levenberg, R.~Monga,
  S.~Moore, D.~G. Murray, B.~Steiner, P.~Tucker, V.~Vasudevan, P.~Warden,
  M.~Wicke, Y.~Yu, and X.~Zheng, ``Tensorflow: A system for large-scale machine
  learning,'' in \emph{12th {USENIX} Symposium on Operating Systems Design and
  Implementation ({OSDI} 16)}.\hskip 1em plus 0.5em minus 0.4em\relax Savannah,
  GA: {USENIX} Association, Nov. 2016, pp. 265--283.

\bibitem{ref:keras}
F.~Chollet \emph{et~al.}, ``Keras,'' \url{https://keras.io}, 2015.

\bibitem{ref:bergstra_hyperopt}
J.~Bergstra, D.~Yamins, and D.~D. Cox, ``Making a science of model search:
  Hyperparameter optimization in hundreds of dimensions for vision
  architectures,'' in \emph{Proceedings of the 30th International Conference on
  International Conference on Machine Learning - Volume 28}, ser. ICML'13,
  2013, p. I–115–I–123.

\bibitem{ref:omalley2019kerastuner}
T.~O'Malley, E.~Bursztein, J.~Long, F.~Chollet, H.~Jin, L.~Invernizzi
  \emph{et~al.}, ``Kerastuner,''
  \url{https://github.com/keras-team/keras-tuner}, 2019.

\bibitem{ref:LeCun_DeepLearning}
Y.~LeCun, Y.~Bengio, and G.~Hinton, ``Deep learning,'' \emph{Nature}, vol. 521,
  pp. 436--444, 2015.

\bibitem{ref:hyperband}
L.~Li, K.~Jamieson, G.~DeSalvo, A.~Rostamizadeh, and A.~Talwalkar, ``Hyperband:
  A novel bandit-based approach to hyperparameter optimization,'' \emph{Journal
  of Machine Learning Research}, vol.~18, no. 185, pp. 1--52, 2018.

\bibitem{ref:brunner}
J.~S. Brunner, M.~A. Makridis, and A.~Kouvelas, ``Comparing the observable
  response times of acc and cacc systems,'' \emph{IEEE Transactions on
  Intelligent Transportation Systems}, pp. 1--10, 2022.

\bibitem{ref:makridis}
M.~Makridis, K.~Mattas, B.~Ciuffo, F.~Re, A.~Kriston, F.~Minarini, and
  G.~Rognelund, ``Empirical study on the properties of adaptive cruise control
  systems and their impact on traffic flow and string stability,''
  \emph{Transportation Research Record}, vol. 2674, no.~4, pp. 471--484, 2020.

\bibitem{gammelli2022graph}
D.~Gammelli, K.~Yang, J.~Harrison, F.~Rodrigues, F.~C. Pereira, and M.~Pavone,
  ``Graph meta-reinforcement learning for transferable autonomous
  mobility-on-demand,'' in \emph{SIGKDD Conference on Knowledge Discovery and
  Data Mining}, 2022.

\end{thebibliography}

\end{document}